\documentclass[sigconf,natbib=true]{acmart}
\usepackage{diagbox}
\usepackage{subfigure}
\usepackage{amsmath}
\usepackage{amsthm}
\usepackage{multirow}
\usepackage{xcolor}
\usepackage{tabularx}
\usepackage{graphicx}
\usepackage{adjustbox}
\usepackage[graphicx]{realboxes}
\usepackage{url}
\usepackage{stfloats}
\usepackage{mathtools}
\usepackage{hyperref}

\usepackage{amssymb,amsfonts}

\usepackage[linesnumbered,ruled,noend]{algorithm2e} 
\DeclareMathOperator*{\argmax}{arg\,max}
\DeclareMathOperator*{\argmin}{arg\,min}
\AtBeginDocument{%
  \providecommand\BibTeX{{%
    \normalfont B\kern-0.5em{\scshape i\kern-0.25em b}\kern-0.8em\TeX}}}

\copyrightyear{2022}
\acmYear{2022}
\setcopyright{acmcopyright}\acmConference[KDD '22]{Proceedings of the 28th ACM SIGKDD Conference on Knowledge Discovery and Data Mining}{August 14--18, 2022}{Washington, DC, USA}
\acmBooktitle{Proceedings of the 28th ACM SIGKDD Conference on Knowledge Discovery and Data Mining (KDD '22), August 14--18, 2022, Washington, DC, USA}
\acmPrice{15.00}
\acmDOI{10.1145/3534678.3539454}
\acmISBN{978-1-4503-9385-0/22/08}

\begin{document}

\title{SOS: Score-based Oversampling for Tabular Data}




\author{Jayoung Kim}
\authornote{Both authors contributed equally to this research.}
\email{jayoung.kim@yonsei.ac.kr}
\orcid{0000-0002-6327-1006}
\author{Chaejeong Lee}
\authornotemark[1]
\email{chaejeong_lee@yonsei.ac.kr}
\affiliation{%
  \institution{Yonsei University}
  \country{South Korea}}

\author{Yehjin Shin}
\email{yehjin.shin@gmail.com}

\affiliation{%
  \institution{Yonsei University}
  \country{South Korea}}

\author{Sewon Park}
\email{sw0413.park@samsung.com}

\affiliation{%
  \institution{Samsung SDS}
  \country{South Korea}}

\author{Minjung Kim}
\email{mj100.kim@samsung.com}

\affiliation{%
 \institution{Samsung SDS}
 \country{South Korea}}

\author{Noseong Park}
\email{noseong@yonsei.ac.kr}

\affiliation{%
  \institution{Yonsei University}
  \country{South Korea}}

\author{Jihoon Cho}
\email{jihoon1.cho@samsung.com}

\affiliation{%
  \institution{Samsung SDS}
  \country{South Korea}}

  \renewcommand{\shortauthors}{Kim and Lee, et al.}
\begin{abstract}
Score-based generative models (SGMs) are a recent breakthrough in generating fake images. SGMs are known to surpass other generative models, e.g., generative adversarial networks (GANs) and variational autoencoders (VAEs). Being inspired by their big success, in this work, we fully customize them for generating fake tabular data. In particular, we are interested in oversampling minor classes since imbalanced classes frequently lead to sub-optimal training outcomes. To our knowledge, we are the first presenting a score-based tabular data oversampling method. Firstly, we re-design our own score network since we have to process tabular data. Secondly, we propose two options for our generation method: the former is equivalent to a style transfer for tabular data and the latter uses the standard generative policy of SGMs. Lastly, we define a fine-tuning method, which further enhances the oversampling quality. In our experiments with 6 datasets and 10 baselines, our method outperforms other oversampling methods in all cases. 

\end{abstract}

\ccsdesc[500]{Computing methodologies~Artificial intelligence}

\keywords{score-based generative model; tabular data synthesis; oversampling}


\maketitle

\section{Introduction}
Tabular data is one of the most frequently occurring data types in the field of data mining and machine learning. However, imbalanced situations happen from time to time, e.g., one class significantly outnumbers other classes. Oversampling minor classes (and as a result, making class sizes even) is a long-standing research topic. Many methods have been proposed, from classical statistical methods~\cite{10.5555/1622407.1622416,10.1007/11538059,4633969}, to recent deep generative methods~\cite{DBLP:journals/corr/abs-1803-09655,Mullick_2019_ICCV}.

\begin{figure}[t]
    \centering
    \subfigure[The overall workflow of score-based generative models, where the score function is approximated by a score network, i.e., $S_{\theta}(\mathbf{x}_t,t) \approx \nabla_\mathbf{x}\log p_t(\mathbf{x})$. We note that it means the gradient of the log probability w.r.t. $\mathbf{x}$ at time $t$.]{\includegraphics[width=0.9\columnwidth]{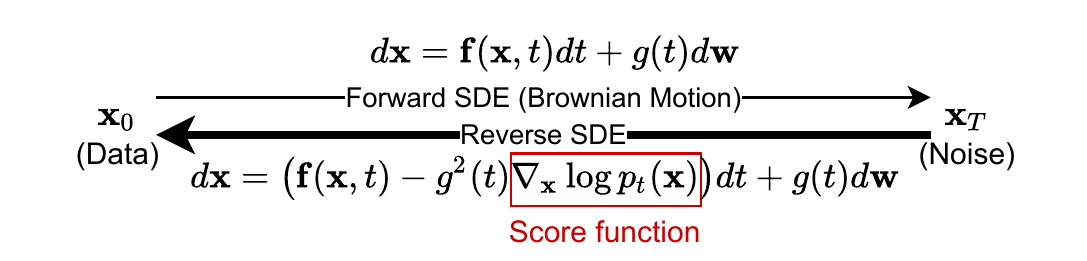}}
    \subfigure[The proposed concept where i) we transfer a non-target (or major) record $\mathbf{x}^+_0$ to a synthesized target (or minor) record $\hat{\mathbf{x}}^-_0$ or ii) we sample a noisy vector $\mathbf{z}$ and generate $\hat{\mathbf{x}}^-_0$.]{\includegraphics[width=0.9\columnwidth]{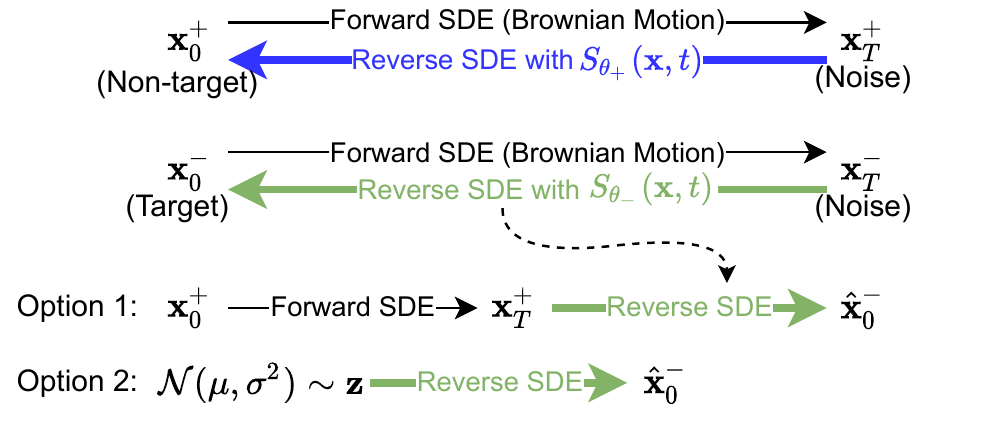}}
    \subfigure[The proposed fine-tuning method where we i) compare the two gradients, and ii) fine-tune $\theta_-$ to enhance the sampling quality.]{\includegraphics[width=0.9\columnwidth]{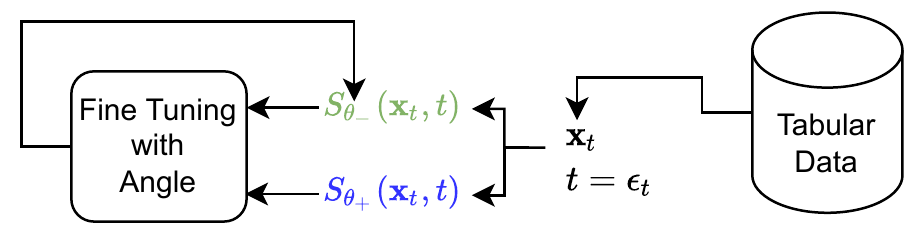}}
     \caption{In this figure, we assume one major (non-target) and one minor (target) classes for ease of discussion but our propose method can be applied to multiple minor classes. (a) The reverse SDE can be considered as a generator since it produces data from noise. (b) The combination of the major (non-target) class's forward SDE and the minor (target) class's reverse SDE can be considered as a \emph{style transfer}-based oversampling process. (c) The fine-tuning process can further enhance the minor (target) class's score network.}
    \label{fig:archi}
    \vspace{-2em}
\end{figure}

Among many deep generative models, score-based generative models (SGMs~\cite{songyang}) have gained much attention recently. They adopt a stochastic differential equation (SDE) based framework to model i) the (forward) diffusion process and ii) the (reverse) denoising process (cf. Fig.~\ref{fig:archi} (a)). The diffusion process is to add increasing noises to the original data $\mathbf{x}_0$, and the denoising process is the reverse of the diffusion process to remove noises from the noisy data $\mathbf{x}_T$. Since the space of $\mathbf{x}$ can be tremendously large, the score function, $\nabla_\mathbf{x}\log p_t(\mathbf{x})$ is approximated by a \emph{score neural network} (or simply a score network), where $t \in [0, T]$ and the step sizes of the diffusion and the denoising processes are typically 1. The ground-truth score values are collected during the forward SDE process. In comparison with generative adversarial networks (GANs), SGMs are known to be superior in terms of generation quality~\cite{songyang,dockhorn2022scorebased,xiao2022tackling}. However, one drawback of SGMs is that they require large-scale computation, compared to other generative models. {The main reason is that SGMs sometimes require long-step SDEs, e.g., $T=1,000$ for images in~\cite{songyang}. It takes long time to conduct the denoising process with such a large number of steps. At the same time, however, it is also known that small $T$ values work well for time series~\cite{rasul2021autoregressive}.}

In this paper, we propose a novel \underline{\textbf{S}}core-based \underline{\textbf{O}}ver\underline{\textbf{S}}ampling (SOS) method for tabular data and found that i) $T$ can be small in our case because of our model design, ii) the score network architecture should be significantly redesigned, and iii) a fine-tuning method, inspired by one characteristic of imbalanced tabular data that we have both target and non-target classes, can further enhance the synthesis quality.

As shown in Fig.~\ref{fig:archi} (b), we separately train two SGMs, one for the non-target (or major) class and the other for the target (or minor) class to oversample --- for simplicity but without loss of generality, we assume the binary class data, but our model is such flexible that multiple minor classes can be augmented. In other words, we separately collect the ground-truth score values of the target and non-target classes and train the score networks. Therefore, their reverse SDEs will be very different. After that, we have two model designs denoted as Option 1 and 2 in Fig.~\ref{fig:archi} (b). By combining the non-target class's forward SDE and the target class's reverse SDE, firstly, we can create a style transfer. A non-target record $\mathbf{x}^+_0$ is converted to a noisy record $\mathbf{x}^+_T$, from which a fake target record $\hat{\mathbf{x}}^-_0$ is created. In other words, the style (or class label) of $\mathbf{x}^+_0$ is transferred to that of the target class. Secondly, we sample a noisy vector $\mathbf{z}$ and generate a fake target record.

Since the forward SDE is a pure Brownian motion process shared by the non-target and the target classes, it constitutes a single noisy space. Since we maintain a single noisy space for both of them, $\mathbf{x}^+_T$ is a correct noisy representation of the non-target record $\mathbf{x}^+_0$ even when it comes to the target class's reverse SDE. Therefore, the first option in Fig.~\ref{fig:archi} (b) can be considered as a style transfer-based oversampling method. Its main motivation is to oversample around class boundaries since the non-target record is (little) perturbed to generate the fake target record (cf. Fig.~\ref{fig:majortominor}).

We further enhance our method by adopting the fine-tuning method in Fig.~\ref{fig:archi} (c), a post-processing procedure after training the two score networks. Our fine-tuning concept is to compare the two gradients (approximated) by $S_{\theta_-}$ and $S_{\theta_+}$ at a certain $(\mathbf{x}_t , t=\epsilon_t)$ pair, where $\mathbf{x} \in \mathcal{T}$ is a real record, $\mathcal{T}$ is tabular data, and $\epsilon_t$ is a time point close to 0 --- note that at this step, we do not care about the class label of $\mathbf{x}$ but use all records in the tabular data. If the angle between the two gradients is small, we regulate $S_{\theta_-}$ to discourage the sampling direction toward the measured gradient (since the direction will lead us to a region where the non-target and target records co-exist).


We conduct experiments with 6 benchmark tabular datasets and 10 baselines: 4 out of 6 datasets are for binary classification and 2 are for multi-class classification. Our careful choices of the baselines include various types of oversampling algorithms, ranging from classical statistical methods, generative adversarial networks (GANs), variational autoencoders (VAEs) to neural ordinary differential equation-based GANs. In those experiments, our proposed method significantly outperforms existing methods by large margins --- in particular, our proposed method is the first one increasing the test score commonly for all those benchmark datasets. Our contributions can be summarized as follows:
\begin{enumerate}
    \item We, for the first time, design a score-based generative model (SGM) for oversampling tabular data. Existing SGMs mainly study image synthesis.
    \item We design a neural network to approximate the score function for tabular data and propose a style transfer-based oversampling technique.
    \item Owing to one characteristic of tabular data that we always have class labels for classification, we design a fine-tuning method, which further enhances the sampling quality.
    \item In our oversampling experiments with 6 benchmark datasets and 10 baselines, our method not only outperforms them but also increases the test score for all cases in comparison with the test score before oversampling (whereas existing methods fail in at least one case).
    \item In Appendix~\ref{a:entire}, our method overwhelms baselines when generating full fake tabular data (instead of oversampling minor classes) as well.
\end{enumerate}

\section{Related Work and Preliminaries}
In this section, we review papers related to our research. We first review the recent score-based generative models, followed by tabular data synthesis.

\subsection{Score-based Generative Models}
SGMs use the following It\^o SDE to define diffusive processes:
\begin{align}\label{eq:forward}
d\mathbf{x}=\mathbf{f}(\mathbf{x},t)dt + g(t)d\mathbf{w},
\end{align}where $\mathbf{f}(\mathbf{x},t) = f(t)\mathbf{x}$ and its reverse SDE is defined as follows:
\begin{align}\label{eq:reverse}
d\mathbf{x}=\big(\mathbf{f}(\mathbf{x},t)-g^2(t)\nabla_\mathbf{x} \log p_t(\mathbf{x})\big)dt + g(t)d\mathbf{w},
\end{align} where this reverse SDE process is a generative process (cf. Fig.~\ref{fig:archi} (a)). According to the types of $f$ and $g$, SGMs can be further classified as i) variance exploding (VE), ii) variance preserving (VP), and ii) sub-variance preserving (Sub-VP) models~\cite{songyang}.

In general, $\mathbf{x}_0$ means a real (image) sample and following the diffusive process in Eq.~\eqref{eq:forward}, we can derive $\mathbf{x}_t$ at time $t \in [0,T]$. These above specific design choices are used to easily approximate the transition probability (or the perturbation kernel) $p(\mathbf{x}_t|\mathbf{x}_0)$ at time $t$ with a Gaussian distribution. The reverse SDE in Eq.~\eqref{eq:reverse} maps a noisy sample at $t=T$ to a data sample at $t=0$.

We can collect the gradient of the log transition probability, $\nabla_{\mathbf{x}_t} \log p(\mathbf{x}_t|\mathbf{x}_0)$, during the forward SDE since it can be analytically calculated. Therefore, we can easily train a score network $S_\theta:\mathbb{R}^{\dim(\mathbf{x})} \times [0,T] \rightarrow \mathbb{R}^{\dim(\mathbf{x})}$ as follows, where $\dim(\mathbf{x})$ means the dimensionality of $\mathbf{x}$, e.g., the number of pixels in the case of images or the number of columns in the case of tabular data:
\begin{align}\label{eq:sgm}
    \argmin_{\theta} \mathbb{E}_t \mathbb{E}_{\mathbf{x}_t} \mathbb{E}_{\mathbf{x}_0} \Big[\lambda(t) \|S_{\theta}(\mathbf{x}_t, t) -\nabla_{\mathbf{x}_t} \log p(\mathbf{x}_t|\mathbf{x}_0) \|_2^2 \Big],
\end{align}where $\lambda(t)$ is to control the trade-off between synthesis quality and likelihood. This training method is known as \emph{denoising score matching} and it is known that $\theta^*$ solving Eq.~\eqref{eq:sgm} can correctly solve the reverse SDE in Eq.~\eqref{eq:reverse}. In other words, the optimal solution $\theta^*$ of the denoising score matching training is equivalent to the optimal solution of the exact score matching training --- its proof is in~\cite{10.1162/NECO_a_00142}.

\begin{figure}
    \centering
    \subfigure[Adversarial score matching where the score network $S_\theta$ is trained with i) the denoising score matching loss in Eq.~\eqref{eq:sgm} and ii) the adversarial training.]{\includegraphics[width=0.9\columnwidth]{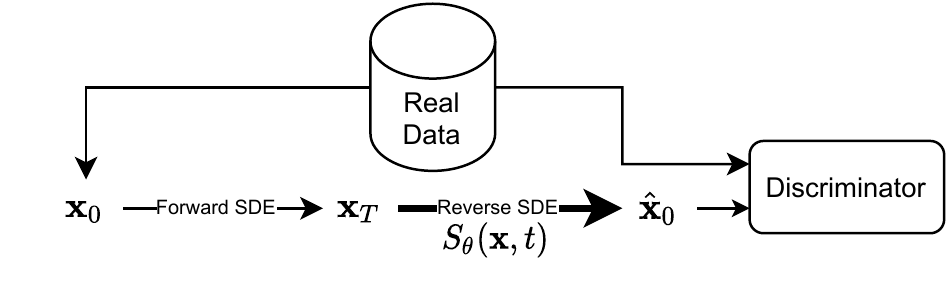}}
    \subfigure[Langevin corrector highlighted in purple]{\includegraphics[width=0.7\columnwidth]{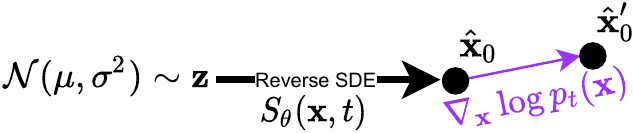}}
    \caption{Two ideas further improving the synthesis quality of SGMs. (a) The adversarial score matching model combining SGMs and GANs. (b) The correction algorithm, called \emph{Langevin corrector}, where $\hat{\mathbf{x}}'_0$ has a higher log probability than $\hat{\mathbf{x}}_0$ since we move following the gradient of the log probability.}
    \label{fig:asm}
    \vspace{-1,5em}
\end{figure}

There exist several other improvements on SGMs. For instance, the adversarial score matching model~\cite{jolicoeur2020adversarial} is to combine SGMs and GANs. Therefore, the reverse SDE is not solely determined as a reverse of the forward SDE but as a trade off between the denoising score matching and the adversarial training of GANs (cf. Fig.~\ref{fig:asm} (a)). 

Once the training process is finished, one can synthesize fake data samples with the \emph{predictor-corrector} framework. In the prediction phase, we solve the reverse SDE in Eq.~\eqref{eq:reverse} after sampling a Gaussian noise. Let $\hat{\mathbf{x}}_0$ be a solution of the reverse SDE. In the correction phase, we then enhance it by using the Langevin corrector~\cite{songyang}. 
The key idea of the correction phase is in Fig.~\ref{fig:asm} (b). Since we correct the predicted sample $\hat{\mathbf{x}}_0$, following the direction of the gradient of the log probability, the corrected sample has a higher log probability. However, only the VE and VP models use this correction mechanism whereas the Sub-VP model does not.


\subsection{Tabular Data Synthesis}

Generative adversarial networks (GANs) are the most popular deep learning-based model for tabular data synthesis, and they generate fake samples by training a generator and a discriminator in an adversarial way~\cite{NIPS2014_5423}. Among many variants of GANs~\cite{pmlr-v70-arjovsky17a,10.5555/3295222.3295327,DBLP:conf/icpr/ParkAMLCP0PK18,NIPS2018_7909,NEURIPS2019_b59a51a3,NIPS2016_7c9d0b1f}, \texttt{WGAN} is widely used providing a more stable training of GANs~\cite{pmlr-v70-arjovsky17a}. Especially, \texttt{WGAN-GP} is one of the most sophisticated model and is defined as follows~\cite{10.5555/3295222.3295327}:
\begin{align*}
\begin{split}
    \mathop{min}_{\substack{\mathtt{G}}}\mathop{max}_{\substack{\mathtt{D}}} \mathbb{E}_{p_{d}}[\mathtt{D}(\mathbf{x})]-\mathbb{E}_{p_{z}}[\mathtt{D}(\mathtt{G}(\mathbf{z}))] -\lambda\mathbb{E}_{p_{\tilde{\mathbf{x}}}}[(\|\nabla_{\tilde{\mathbf{x}}}\mathtt{D}(\tilde{\mathbf{x}})\|_2-1)^2],
\end{split}\end{align*} where $\mathtt{G}$ is a generator function, $\mathtt{D}$ is a Wasserstein critic function, $p_{\mathbf{z}}$ is a prior distribution for a noisy vector $\mathbf{z}$, which is typically $\mathcal{N}(\mathbf{0},\mathbf{1})$, $p_{d}$ is a data distribution, $\tilde{\mathbf{x}}$ is a randomly weighted combination of $\mathtt{G}(\mathbf{z})$ and $\mathbf{x}$, and $\lambda$ is a regularization weight. 

Tabular data synthesis generates realistic table records by modelling a joint probability distribution of features in a table. \texttt{CLBN}~\cite{1054142} is a Bayesian network built by the Chow-Liu algorithm representing a joint probability distribution. \texttt{TVAE}~\cite{NIPS2019_8953} is a variational autoencoder that effectively handles mixed types of features in tabular data. There are also many variants of GANs for tabular data synthesis. \texttt{RGAN}~\cite{esteban2017realvalued} generates continuous time-series records while \texttt{MedGAN}~\cite{DBLP:journals/corr/ChoiBMDSS17} generates discrete medical records using non-adversarial loss. \texttt{TableGAN}~\cite{DBLP:journals/corr/abs-1806-03384} generates tabular data using convolutional neural networks.
\texttt{PATE-GAN}~\cite{Jordon2019PATEGANGS} was proposed to synthesize differentially private tabular records.

However, one notorious challenge in synthesizing tabular data with GANs is addressing the issue of mode collapse. In many cases, the fake data distribution synthesized by GANs is confined to only a few modes and fails to represent many others. Some variants of GANs tried to resolve this problem, proposing to force the generator to produce samples from the various modes as evenly as possible. \texttt{VEEGAN}~\cite{NIPS2017_6923} has a reconstructor network, which maps the output of the generator to the noisy vector space, and jointly trains the generator and the reconstructor network. \texttt{CTGAN}~\cite{NIPS2019_8953} has a conditional generator and use a mode separation process. \texttt{OCT-GAN}~\cite{10.1145/3442381.3449999} exploits the homeomorphic characteristic of neural ordinary differential equations, when designing its generator, and now shows the state-of-the-art synthesis quality for many tabular datasets. However, one downside is that it requires a much longer training time than other models.

\begin{table}
\small
\centering
\setlength{\tabcolsep}{3pt}
\caption{The qualitative comparison among our method and other methods. There are no existing minor-class oversampling models that adopt the score-based generative model. Moreover, many of the existing minor-class oversampling methods are for images. In our experiments, we modify some of them to be able to process tabular data and compare with our method.}\label{tbl:cmp}
\begin{tabular}{c|c|c|c}
\specialrule{1pt}{1pt}{1pt}

Method & Multiple Generators & Generative Model Type & Domain \\ 
\specialrule{1pt}{1pt}{1pt}
\texttt{SMOTE} & N/A & Classical Method & Tabular \\ \hline
\texttt{B-SMOTE} & N/A & Classical Method & Tabular \\ \hline
\texttt{ADASYN} & N/A & Classical Method & Tabular \\ \hline
\texttt{BAGAN} & X & GAN & Image \\ \hline
\texttt{cWGAN} & X & GAN & Tabular \\ \hline
\texttt{GL-GAN} & X & GAN & Tabular \\ \hline
\texttt{GAMO} & O & GAN & Image \\ \hline
\texttt{SOS} (Ours) & O (Score Networks) & Score-based Model & Tabular \\  
\specialrule{1pt}{1pt}{1pt}

\end{tabular}
\end{table}

\subsection{Minor-class Oversampling}
Multi-class data commonly suffers from imbalanced classes, and two different approaches have been proposed to address the issue; classical statistical methods and deep learning-based methods.

Classical statistical methods are mostly based on the synthetic minority oversampling technique (\texttt{SMOTE}~\cite{10.5555/1622407.1622416}) method, which generates synthetic samples by interpolating two minor samples. Borderline SMOTE (\texttt{B-SMOTE}~\cite{10.1007/11538059}) improves \texttt{SMOTE} to consider only the minor samples on the class boundary where the major samples occupy more than half of the neighbors. The adaptive synthetic sampling (\texttt{ADASYN}~\cite{4633969}) applies \texttt{SMOTE} on the minority samples close to the major samples. 

Deep learning-based methods, based on GANs, mostly aim at oversampling minor class in images. In \texttt{ACGAN}~\cite{odena2017conditional}, a generator generates all classes including the minor ones, and a discriminator returns not only whether the input is fake or real but also which class it belongs to. Inspired by \texttt{ACGAN}, \texttt{BAGAN}~\cite{DBLP:journals/corr/abs-1803-09655} constructs the discriminator which distinguishes the fake from the rest of the classes at once, showing good performance for minor-class image generation. \texttt{cWGAN}~\cite{Engelmann2021ConditionalWG} is a WGAN-GP based tabular data oversampling model, which is equipped with an auxiliary classifier to consider the downstream classification task. \texttt{GL-GAN}~\cite{wang2020global} is a GAN that employs an autoencoder and SMOTE to learn and oversample the latent vectors and includes an auxiliary classifier to provide label information to the networks. \texttt{GAMO}~\cite{Mullick_2019_ICCV} uses multiple generators, each of which is responsible for generating a specific class of data, and two separate critics; one for the real/fake classification and another for the entire class classification.

\section{Proposed Method}
We describe our proposed score-based oversampling method. We first sketch the overall workflow of our method, followed by its detailed designs. 

\begin{figure}
    \centering
    \subfigure[Option 1: Style transfer-based oversampling which corresponds to \emph{oversampling around class boundary}]{\includegraphics[width=\columnwidth]{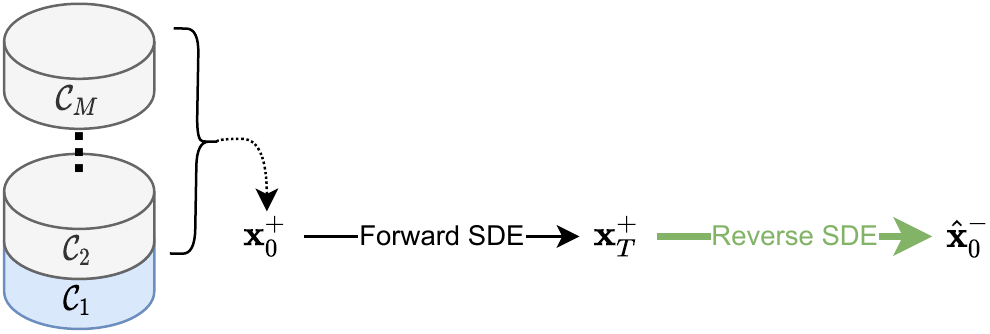}}
    \subfigure[Option 2: Plain score-based oversampling]{\includegraphics[width=0.6\columnwidth]{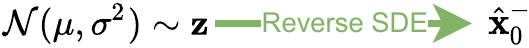}}
    \caption{Another diagram showing our model architectures. In this figure, we assume multiple minor classes. Given a set of classes $\{\mathcal{C}_m\}_{m=1}^M$, suppose that a target minor class is $\mathcal{C}_1$, for which we aim at oversampling. (a) We randomly choose $\mathbf{x}^+_0$ from other non-target classes $\{\mathcal{C}_m\}_{m=2}^M$. (b) We randomly sample a noisy vector $\mathbf{z}$.}
    \label{fig:patrick}
    \vspace{-1em}
\end{figure}

\begin{definition}
Let $\mathcal{T}$ be a table (or tabular data) consisting of $M$ classes, $\{\mathcal{C}_m\}_{m=1}^M$, where $\mathcal{C}_m$ is $m$-th class (subset) of $\mathcal{T}$, i.e., classes are disjoint subsets of $\mathcal{T}$. We define a minor class as $\mathcal{C}_j$ whose cardinality is smaller than the maximum cardinality, i.e., $|\mathcal{C}_j| < \max_m |\mathcal{C}_m|$. In general, therefore, we have one major (with the largest cardinality) and multiple minor classes in $\mathcal{T}$. Our task is to oversample those minor classes, i.e., add artificial fake minor records, until $|\mathcal{C}_j| = \max_m |\mathcal{C}_m|$ for all $j$.
\end{definition}

\subsection{Overall Workflow}\label{workflow}

Our proposed score-based oversampling, shown in Fig.~\ref{fig:patrick}, is done by the following workflow:
\begin{enumerate}
    \item We train a score-based generative model for each class $\mathcal{C}_m$. Let $S_{\theta_m}$ be a trained score network for $\mathcal{C}_m$.
    \item Let $\mathcal{C}_j$ be a minor class to oversample, i.e., target class. We fine-tune $S_{\theta_j}$ using all records of $\mathcal{T}$ as shown in Fig.~\ref{fig:archi} (c).
    \item In order to oversample each target (minor) class $\mathcal{C}_j$, we use one of the following two options as shown in Fig.~\ref{fig:patrick}. We repeat this until $|\mathcal{C}_j| = \max_m |\mathcal{C}_m|$.
    \begin{enumerate}
        \item After feeding a record randomly chosen from all other non-target classes except $\mathcal{C}_j$, we derive its noisy vector. By using $S_{\theta_j}$, we synthesize a fake target (minor) record for $\mathcal{C}_j$. (cf. Fig.~\ref{fig:patrick} (a)).
        \item After sampling a noisy vector $\mathbf{z} \sim \mathcal{N}(\mu, \sigma^2)$, we generate a fake target (minor) record (cf. Fig.~\ref{fig:patrick} (b)).
    \end{enumerate}
    \item After generating a fake record for the target class $\mathcal{C}_j$, we can apply the Langevin corrector, as noted in Fig.~\ref{fig:asm} (b). However, this step is optional and for some types of SGMs, we also do not use this step sometimes --- see Appendix~\ref{a:hyper} for its detailed settings in each dataset for our experiments. For instance, the type of Sub-VP does not use this step for reducing computation in~\cite{songyang} since it shows good enough quality even without the corrector.
\end{enumerate}

\begin{definition}
Let $\mathcal{C}_j$ be a target minor class, for which we aim at oversampling. $\mathbf{x}^-_0 \in \mathcal{C}_j$ is a target (minor) record in the target class.
\end{definition}

\begin{definition}
Let $\mathcal{C}_j$ be a target minor class, for which we aim at oversampling. $\mathbf{x}^+_0 \in \mathcal{C}_i$, where $i \neq j$, is a record in other non-targeted classes. In the case of binary classification, $\mathcal{C}_j$ is a minor class and $\mathcal{C}_i$ is a major class. In general, however, $\mathbf{x}^+_0$ means a record from other non-target classes. For ease of our convenience for writing, we use the symbol `$+$' to denote all other non-targeted classes. Fig.~\ref{fig:patrick} follows this notation.
\end{definition}

In this process, our main contributions lie in i) designing the score network $S_{\theta_m}$ for each class $\mathcal{C}_m$, and ii) designing the fine-tuning method. Note that the fine-tuning happens after all score networks are trained.

\paragraph{\textbf{Option 1 vs. Option 2}} In Fig.~\ref{fig:patrick}, we show two options of how to sample a noisy vector. In Fig.~\ref{fig:patrick} (a), we perturb $\mathbf{x}^+_0$ in a non-target class to generate $\hat{\mathbf{x}}^-_0$. This strategy can be considered as \emph{oversampling around class boundary}. In Fig.~\ref{fig:patrick} (b), we sample from random noisy vectors and therefore, there is no guarantee that $\hat{\mathbf{x}}^-_0$ is around class boundary. We refer to Fig.~\ref{fig:majortominor} for its intuitive visualization result.

{One more advantage of the first option is that we do not require that $\mathbf{x}^+_T$ follows a Gaussian distribution. Regardless of the distribution of $\mathbf{x}^+_T$, we can derive $\hat{\mathbf{x}}^-_0$ using the reverse SDE with $S_{\theta_-}$ --- recall that we do not sample the noisy vector from a Gaussian in the first option but from $\mathbf{x}^+_0$ using the forward SDE. Therefore, we can freely choose a value for $T$, which is one big advantage in our design since a large $T$ increases computational overheads. For the second option, we also found that small $T$ settings work well as in~\cite{rasul2021autoregressive}.

The relationship between the first and the second option is analogous to that between {\texttt{Borderline-SMOTE}} (or \texttt{B-SMOTE}) and \texttt{SMOTE}. In this work, we propose those two mechanisms for increasing the usability of our method.
}

\subsection{Score-based Model for Tabular Data}
We use the forward and reverse SDEs in Eqs.~\eqref{eq:forward} and~\eqref{eq:reverse}. As noted earlier, there exist three types of SGMs, depending on the types of $f$ and $g$ in Eqs.~\eqref{eq:forward} and~\eqref{eq:reverse}. We introduce the definitions of $f$ and $g$ in Appendix~\ref{sec:fandg} and the proposed architecture for our score network in Appendix~\ref{a:archi}.

Since the original score network design is for images, we redesign it with various techniques used by tabular data synthesis models~\cite{NIPS2019_8953,lee2021invertible, grathwohl2018ffjord}. Its details are in Appendix~\ref{a:archi}. However, we train this score network using the denoising score matching loss as in the original design.


Our method provides two options for oversampling $\mathcal{C}_j$. In the first option, we adopt the style transfer-based architecture, where i) we choose $\mathbf{x}^+_0$ from $\mathcal{C}_i$, $i \neq j$, ii) we derive a noisy vector from it using the forward SDE, and iii) we finally transfer the noisy vector to a fake target record. The rationale behind this design is that the forward SDE is a pure Brownian motion regardless of classes --- in other words, all classes share the same forward SDE and thereby, share the same noisy space as well. Therefore, $\mathbf{x}^+_T$ contains small information on its original record $\mathbf{x}^+_0 \in \mathcal{C}_i$, which is still effective when it comes to generating a fake target record for $\mathcal{C}_j$. Thus, this option can be considered as oversampling around class boundary.

In the second option, we follow the standard use of SGMs, where we sample $\mathbf{z} \sim \mathcal{N}(\mu, \sigma^2)$ and generate a fake target record. We note that the Gaussian distribution $\mathcal{N}(\mu, \sigma^2)$ is defined by the types of SDEs, i.e., VE, VP, or Sub-VP, as follows:
\begin{align}
    \mathbf{z} \sim \begin{cases}
     \mathcal{N}(\mathbf{0}, \sigma^2_{max}), &\textrm{ if VE,}\\
     \mathcal{N}(\mathbf{0}, \mathbf{1}), &\textrm{ if VP,}\\
     \mathcal{N}(\mathbf{0}, \mathbf{1}), &\textrm{ if Sub-VP,}\\
    \end{cases}
\end{align}
where $\sigma_{max}$ is a hyperparameter.

\begin{figure}
    \centering
    \includegraphics[width=0.7\columnwidth]{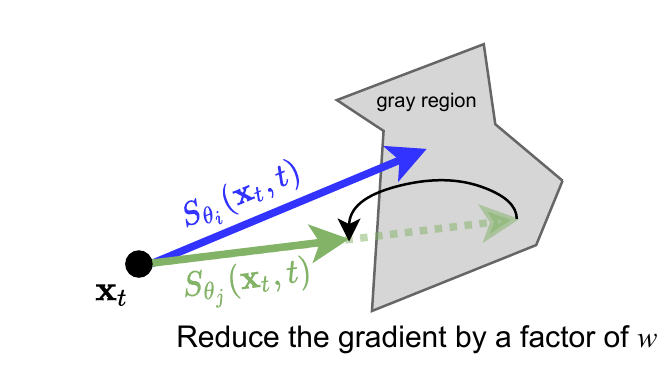}
    \caption{Suppose that we want to fine-tune $\theta_j$ for class $\mathcal{C}_j$. At $(\mathbf{x}_t,t)$, the two gradients evaluated with Eq.~\eqref{eq:eval}, $\mathbf{g}^j_{\mathbf{x}, t}$ and $\mathbf{g}^i_{\mathbf{x}, t}$, create a small angle. Then, we suppress the denoising process (or reverse SDE) toward $\mathbf{g}^j_{\mathbf{x}, t}$ by a factor of $w$, as noted in Eq.~\eqref{eq:tune}, to avoid the gray region where both target and non-target class records are likely to {co-exist}.}
    \label{fig:tune}
\vspace{-1em}
\end{figure}

\subsection{Fine-tuning Method}
Once score networks $\{S_{\theta_m}\}_{m=1}^M$ are trained, we can sample fake records for all target classes as described earlier. In the case of tabular data, however, we can exploit class information to further enhance the sampling quality. Our proposed fine-tuning procedures are as follows --- for simplicity, we describe how to fine-tune for $\mathcal{C}_j$ only. However, we repeat the following method for each $j$:
\begin{enumerate}
\item Firstly, given a record $\mathbf{x} \in \mathcal{T}$ regardless of its class label and a chosen time $t=\epsilon_t$\footnote{$\epsilon_t$ is a small value close to 0, and the range $[0,\epsilon_t]$ means the last moment of the reverse SDE in Eq.~\eqref{eq:reverse}, which we will fine-tune.}, we evaluate with each score network $S_{\theta_m}$ to know the gradient of the log probability w.r.t. $\mathbf{x}_t$ at time $t$, i.e., time-dependant score, as follows:
\begin{align}\label{eq:eval}
    \mathbf{g}^m_{\mathbf{x}, t} = S_{\theta_m}\bigg\rvert_{\mathbf{x}_t, t}, \forall m,
\end{align}where $\mathbf{g}^m_{\mathbf{x}, t}$ is the gradient evaluated at $(\mathbf{x}_t, t)$ with the score network $S_{\theta_m}$. The long vertical bar means the `evaluated-at' operator.
\begin{enumerate}
    \item Our goal is to fine-tune the model around $(\mathbf{x}_t, t)$. Therefore, we add a random noisy vector to $\mathbf{x}$.
\end{enumerate}


\item Secondly, we compare the angle between $\mathbf{g}^j_{\mathbf{x}, t}$ and $\mathbf{g}^i_{\mathbf{x}, t}$ where $i \neq j$. Suppose that the angle $arccos\Big(\frac{\mathbf{g}^j_{\mathbf{x}, t} \cdot \mathbf{g}^i_{\mathbf{x}, t}}{\|\mathbf{g}^j_{\mathbf{x}, t}\| \|\mathbf{g}^i_{\mathbf{x}, t}\|}\Big)$ is large enough. It then says that $\mathcal{C}_j$ and $\mathcal{C}_i$ can be well separated around the location and time of $(\mathbf{x}, t)$, which is a preferred situation. However, a problematic situation happens when the angle is smaller than a threshold {$0 < \xi < \pi$} --- in other words, their directions are similar. In such a case, we use the following loss to fine-tune the gradient $S_{\theta_j}$ at $(\mathbf{x}_t, t)$:
\begin{align}\label{eq:tune}
L(\mathbf{x}, t, j) = \| S_{\theta_j}(\mathbf{x}_t, t) - w \mathbf{g}^j_{\mathbf{x}, t}\|_2^2,
\end{align}where $0 < w < 1$. In other words, we want to decrease the gradient by a factor of $w$.
\end{enumerate}

The rationale behind the proposed fine-tuning method is that by decreasing the scale of the gradient by a factor of $w$ at $(\mathbf{x}_t, t)$, we can prevent that the reverse SDE process moves too much toward the \emph{gray region} where both target and non-target class records are mixed (cf. Fig.~\ref{fig:tune}). By controlling the damping coefficient $w$, we control how aggressively we suppress the direction.

\setlength{\textfloatsep}{8pt}
\begin{algorithm}[t]
\footnotesize
\DontPrintSemicolon
  \caption{How to train our proposed model}\label{alg1}

  Initialize $\theta_m$ for all $1 \leq m \leq M$\;
  
  \For{each class $\mathcal{C}_m$}{
        Train its score network $S_{\theta_m}$ using the denoising score matching loss in Eq.~\eqref{eq:sgm}\;\label{alg:s1}
  }
  
  \tcc{Let $\mathcal{C}_i$ be the major class, where $i=\argmax_m |\mathcal{C}_m|$.}
  \For{each record $\mathbf{x} \in \mathcal{T}$}{
    \For{each minor class $\mathcal{C}_j$}{
        \If{$arccos\Big(\frac{\mathbf{g}^j_{\mathbf{x}, t} \cdot \mathbf{g}^i_{\mathbf{x}, t}}{\|\mathbf{g}^j_{\mathbf{x}, t}\| \|\mathbf{g}^i_{\mathbf{x}, t}\|}\Big) < \xi$}{
          Fine-tune $\theta_j$ with $L(\mathbf{x}, t, j)$ in Eq.~\eqref{eq:tune}\;\label{alg:s2}
      }  
    }
  }
  
  \For{each minor class $\mathcal{C}_j$}{
    Oversample fake minor records with Option 1 or 2 for $\mathcal{C}_j$ until $|\mathcal{C}_j| = \max_m |\mathcal{C}_m|$\;
  }
  
  \Return oversampled minor classes
\end{algorithm}

\subsection{Training Algorithm}

Algorithm~\ref{alg1} shows the overall training process for our method. At the beginning, we initialize the score network parameters for all $m$. We then train each score network $S_{\theta_m}$ using each $\mathcal{C}_m$ in Line~\ref{alg:s1}. At this step, we use the standard denoising score matching loss in Eq.~\eqref{eq:sgm}. After finishing this step, we can already generate fake target records. However, we further enhance each score network in Line~\ref{alg:s2}. The complexity of the proposed fine-tuning is  $\mathcal{O}(|\mathcal{T}|m)$, which is affordable in {practice} since $m$, the number of classes, is not large in comparison with the table size $|\mathcal{T}|$. For oversampling $\mathcal{C}_j$, we can use either the first or the second option of our model designs in Fig.~\ref{fig:patrick}. They correspond to i) oversampling around class boundary or ii) regular oversampling, respectively.

\section{Experiments}
We describe the experimental environments and results. In Appendix~\ref{a:entire}, we introduce our additional experiments that we generate entire fake tabular data (instead of making imbalanced tabular data balanced after oversampling minor classes). We also analyze the space and time overheads of our method in Appendix~\ref{a:comp}.

\vspace{-0.5em}
\subsection{Experimental Environments}
\subsubsection{Datasets} 
We describe the tabular datasets that are used for our experiments. \texttt{Buddy} and \texttt{Satimage} are for multi-class classification, and the others for binary classification. 

\texttt{Default}~\cite{default} is a dataset describing the information on credit card clients in Taiwan regarding default payments, demographic factors, credit data, and others. {The `\textit{default payment}' column includes 23,364 (77.9$\%$) of major records and 6,636 (22.1$\%$) minor records.} \texttt{Shoppers}~\cite{shoppers} is a binary classification dataset, where out of 12,330 sessions, 10,422 (84.5$\%$) are negative records which did not end with purchase, and the rest 1,908 (15.5$\%$) are positive records ending with purchase. \texttt{Surgical}~\cite{surgical} is a binary classification dataset, which contains general surgical patient information. {There are 10,945 (74.8$\%$) records for the major class, and 3,690 (25.2$\%$) records for the minor class.} \texttt{WeatherAUS}~\cite{weatheraus} contains 10 years of daily weather observations from several weather stations in Australia. {`\textit{RainTomorrow}' is the target variable to classify, where 43,993 (78.0$\%$) records belong to the major class, which means the rain for that day was over 1mm, and 12,427 (22.0$\%$) records belong to the minor class.}

\texttt{Buddy}~\cite{buddy} consists of parameters, including a unique ID, assigned to each animal that is up for adoption, and a variety of physical attributes. {The target variable is `\textit{breed}', which includes 10,643 (56.5$\%$) records for the major class, and 7,021 (37.3$\%$) and 1,170 (6.2$\%$) records for minor classes.} \texttt{Satimage}~\cite{satimage} is to classify types of land usage, which is created by the Australian Centre for Remote Sensing. {The dataset contains 2,282 (35.8$\%$) records for the major class, and 1,336 (21.0$\%$), 1,080 (16.9$\%$), 562 (8.8$\%$), 573 (9.0$\%$), and 539 (8.5$\%$) records for minor classes.}

\vspace{-0.3em}
\subsubsection{Baselines}

We use a set of baselines as follows, which include statistical methods and generative models based on deep learning:

\texttt{Identity} is the original tabular data without oversampling. 
\texttt{SMOTE}~\cite{10.5555/1622407.1622416}, \texttt{B-SMOTE}~\cite{10.1007/11538059}, and \texttt{Adasyn}~\cite{4633969} are classical methods of data augmentation for the minority class;
\texttt{MedGAN}~\cite{DBLP:journals/corr/ChoiBMDSS17}, \texttt{VEEGAN}~\cite{NIPS2017_6923}, \texttt{TableGAN}~\cite{DBLP:journals/corr/abs-1806-03384}, and \texttt{CTGAN}~\cite{NIPS2019_8953} are GAN models for tabular data synthesis;
\texttt{TVAE}~\cite{NIPS2019_8953} is proposed to learn latent embedding in VAE by incorporating deep learning metric. 
\texttt{BAGAN}~\cite{DBLP:journals/corr/abs-1803-09655} is a type of GAN model for oversampling images, but we replace its generator and discriminator with ours for tabular data synthesis.
\texttt{OCT-GAN}~\cite{10.1145/3442381.3449999} is one of the state of the art model for tabular data systhesis.

    

We exclude \texttt{cWGAN}~\cite{Engelmann2021ConditionalWG} and \texttt{GL-GAN}~\cite{wang2020global}, since they do not support multiple minority cases and synthesis quality is also relatively lower than others in our preliminary experiments. 
\vspace{-0.3em}
\subsubsection{Hyperparameters}
We refer to Appendix~\ref{a:hyper} for detailed hyperparameter settings. We also provide all our codes and data for reproducibility and one can easily reproduce our results.
\vspace{-0.3em}
\subsubsection{Evaluation Methods} Let $\mathcal{C}_j$ be a minor class, where $|\mathcal{C}_j| < \max_m |\mathcal{C}_m|$. As mentioned earlier, our task is to oversample  $\mathcal{C}_j$ until $|\mathcal{C}_j| = \max_m |\mathcal{C}_m|$ for all $j$. However, some baselines are not for oversampling minor classes but generating one entire fake tabular data, in which case we train such a baselines for each $\mathcal{C}_j$ --- in other words, one dedicated baseline model for each minor class. For our method, we follow the described process.

For evaluation, we first train various classification algorithms with the augmented (or oversampled) training data --- we note that the augmented data is fully balanced after oversampling --- including \texttt{Decision Tree}~\cite{breiman1984classification}, \texttt{Logistic Regression}~\cite{cox1958regression}, \texttt{AdaBoost}~\cite{10.5555/1624312.1624417}, and \texttt{MLP}~\cite{10.5555/1162264}. We then conduct classification with the testing data. We take the average of the results  --- the same evaluation scenario had been used in~\cite{NIPS2019_8953,lee2021invertible} and we strictly follow their evaluation protocols. We repeat with 5 different seed numbers and report their mean and std. dev. scores. We use the weighted F1 since the testing data is also imbalanced. We visualize the column-wise histogram of original and fake records and use t-SNE~\cite{JMLR:v9:vandermaaten08a} to display them.

\begin{table*}[t]

\caption{Experimental results (Weighted F1).
The best results are in boldface --- if tie, {the winner is determined by comparing in the 4th decimal place.}}
\label{table:main}
\centering
\setlength{\tabcolsep}{5pt}

\begin{tabular*}{0.85\textwidth}{c|c|cccccccccccc}

\specialrule{1pt}{1pt}{1pt}
&\multirow{2}{*}{Methods}  & & & \multicolumn{4}{c}{Single Minority}                                            &  &  & \multicolumn{2}{c}{Multiple Minority}           & &      \\ \cline{5-8} \cline{11-12}
                        & & &  & \texttt{Default} & \texttt{Shoppers} & \texttt{Surgical} & \texttt{WeatherAUS} &  &  & \texttt{Buddy} & \texttt{Satimage}                 &  &  \\ \specialrule{1pt}{1pt}{1pt}
\multicolumn{2}{c|}{\texttt{Identity}}         &&  & 0.515\footnotesize{±0.035}    & 0.601\footnotesize{±0.039}     & 0.687\footnotesize{±0.004}     & 0.657\footnotesize{±0.016}       &  &  & 0.603\footnotesize{±0.010}          & 0.817\footnotesize{±0.004}       &         \\ \specialrule{1pt}{1pt}{1pt}
\multirow{10}{*}{\rotatebox[origin=c]{90}{Baselines}} & \texttt{SMOTE}           & &  & 0.561\footnotesize{±0.025}    & 0.648\footnotesize{±0.004}    & 0.678\footnotesize{±0.008}     & 0.674\footnotesize{±0.025}     &  &  & 0.584\footnotesize{±0.005}     & 0.846\footnotesize{±0.005}   & &           \\
&\texttt{B-SMOTE}         & &  & 0.561\footnotesize{±0.029}    & 0.640\footnotesize{±0.042}    & 0.671\footnotesize{±0.004}    & 0.663\footnotesize{±0.022}    &  &  & 0.595\footnotesize{±0.003}    & 0.845\footnotesize{±0.005}    & &          \\
&\texttt{Adasyn}          & &  & 0.558\footnotesize{±0.023}    & 0.630\footnotesize{±0.045}    & 0.662\footnotesize{±0.007}    & 0.658\footnotesize{±0.022}    &  &  & 0.608\footnotesize{±0.002}           & 0.841\footnotesize{±0.008}     &  &        \\
&\texttt{MedGAN}          & &  & 0.532\footnotesize{±0.028}    & 0.620\footnotesize{±0.062}    & 0.686\footnotesize{±0.003}    & 0.656\footnotesize{±0.022}      &  &  & 0.598\footnotesize{±0.011}     & 0.835\footnotesize{±0.019}  &  &        \\
&\texttt{VEEGAN}          & &  & 0.495\footnotesize{±0.076}    & 0.607\footnotesize{±0.065}   & 0.680\footnotesize{±0.117} & 0.661\footnotesize{±0.025}    &  &  & 0.555\footnotesize{±0.036}           & 0.840\footnotesize{±0.031} &   &       \\
&\texttt{TableGAN}        & &  & 0.423\footnotesize{±0.115}    & 0.571\footnotesize{±0.097}     & 0.704\footnotesize{±0.001} & 0.579\footnotesize{±0.066}       &  &  & 0.570\footnotesize{±0.019}           & 0.813\footnotesize{±0.013}      &    &      \\
&\texttt{TVAE}            & &  & 0.536\footnotesize{±0.035}    & 0.610\footnotesize{±0.060}    & 0.681\footnotesize{±0.004}    & 0.652\footnotesize{±0.018}       &  &  & 0.552\footnotesize{±0.044}           & 0.846\footnotesize{±0.031}     &     &     \\
&\texttt{CTGAN}           & &  & 0.545\footnotesize{±0.022}    & 0.605\footnotesize{±0.059}    & 0.701\footnotesize{±0.004}    & 0.659\footnotesize{±0.020}      &  &  & 0.593\footnotesize{±0.009}           & 0.833\footnotesize{±0.015}      &      &    \\
&\texttt{OCT-GAN}          &&  & 0.531\footnotesize{±0.018}    & 0.639\footnotesize{±0.029}    & 0.692\footnotesize{±0.082}    & 0.656\footnotesize{±0.018}       &  &  & 0.551\footnotesize{±0.015}           & 0.837\footnotesize{±0.011}      &       &   \\ 
&\texttt{BAGAN}           & &  & 0.525\footnotesize{±0.005}    & 0.610\footnotesize{±0.005}    & 0.668\footnotesize{±0.004}    & 0.663\footnotesize{±0.002}       &  &  & 0.555\footnotesize{±0.013}           & 0.834\footnotesize{±0.011}      &        &  \\ \hline
\multirow{3}{*}{\rotatebox[origin=c]{90}{\texttt{SOS}}} & VE         & &  &  0.571\footnotesize{±0.003} &   \textbf{0.675\footnotesize{±0.004}}   &  0.709\footnotesize{±0.003}  &   0.672\footnotesize{±0.002}    &  &  &    0.607\footnotesize{±0.007}       &      0.854\footnotesize{±0.002}     &      &\\
& VP         & &  &  0.559\footnotesize{±0.006} & 0.658\footnotesize{±0.003}  &  0.712\footnotesize{±0.002} &   0.680\footnotesize{±0.002}     &  &  &     0.607\footnotesize{±0.011}     &         \textbf{0.857\footnotesize{±0.006}}   & &    \\ 
& Sub-VP      & &  & \textbf{0.574\footnotesize{±0.003}}  &  0.673\footnotesize{±0.002}    & \textbf{0.714\footnotesize{±0.001}}     & \textbf{0.680\footnotesize{±0.003}}      &  &  & \textbf{0.608\footnotesize{±0.002}}       & 0.855\footnotesize{±0.004}      &  &      \\ 
\specialrule{1pt}{1pt}{1pt}

\end{tabular*}   
\vspace{-0.5em}
\end{table*}

\vspace{-0.3em}
\subsection{Experimental Results}
We compare various oversampling methods in Table~\ref{table:main}. As mentioned earlier, \texttt{Identity} means that we do not oversampling minor classes, train classifiers, and test with testing data. Therefore, the score of \texttt{Identity} can be considered as a minimum requirement upon oversampling.

Classical methods, such as \texttt{SMOTE}, \texttt{B-SMOTE}, and \texttt{Adasyn}, show their effectiveness to some degree. For instance, \texttt{SMOTE} improves the test score from {0.515} to {0.561} after oversampling in \texttt{Default}. All these methods, however, fail to increase the test score in \texttt{Surgical}. \texttt{SMOTE} and \texttt{B-SMOTE} fail in \texttt{Buddy} as well.

\texttt{VEEGAN} and \texttt{TableGAN} are two early GAN-based methods and their test scores are relatively lower than other advanced methods, such as \texttt{CTGAN} and \texttt{OCT-GAN}. However, \texttt{TableGAN} shows the best result among all baselines for \texttt{Surgical}. \texttt{MedGAN} fails in \texttt{Surgical}, \texttt{WeatherAUS}, and \texttt{Buddy}. \texttt{TVAE} also fails in multiple cases. Among the baselines, \texttt{CTGAN} and \texttt{OCT-GAN} show the best outcomes in many cases and in general, their synthesis quality looks stable. In general, those deep learning-based methods show better effectiveness than the classical methods.

However, our method, \texttt{SOS}, clearly outperforms all those methods in all cases by large margins. In \texttt{Default}, the original score without oversampling, i.e., \texttt{Identity}, is 0.515, {and the best baseline score is 0.561 by \texttt{SMOTE}. However, our method with Sub-VP achieves 0.574, which is about 11\%} improvement over \texttt{Identity}. The biggest enhancement is achieved in \texttt{Shoppers}, i.e., a score of 0.601 by \texttt{Identity} vs. 0.675 by our method with VE. For other datasets, our method marks significant enhancements after oversampling. 
In \texttt{Buddy}, moreover, other methods except \texttt{Adasyn} and \texttt{SOS} oversample noisy fake records and decrease the score below that of \texttt{Identity}.


\vspace{-1em}
\subsection{Sensitivity on Hyperparameters}
We summarize the sensitivity results w.r.t. some key hyperparameters in Table~\ref{table:sensitivity}. We used $T=50$ in Table~\ref{table:main}. Others for $T$ produce sub-optimal results in Table~\ref{table:sensitivity}. $\beta_{min}$ and $\beta_{max}$ means the lower and upper bounds of $\beta(t)$ of the drift $f$ and the diffusion $g$. In general, all settings produce reasonable outcomes, outperforming \texttt{Identity}. $\xi$ is the angle threshold for our fine-tuning method. As shown in Appendix~\ref{a:hyper}, the best setting for $\xi$ varies from one dataset to another but in general, $\xi \in [80,100]$ produces the best fine-tuning results. For $w$, we recommend a value close to 0.99 as shown in Table~\ref{table:sensitivity}. $\epsilon_{t}=5e-04$ produces optimal results in almost all cases.

\begin{table}[]
\caption{Results by some selected key hyperparameters}
\label{table:sensitivity}
\small
\centering
\setlength{\tabcolsep}{3pt}

\begin{tabular}{c|cccccccccc}
\specialrule{1pt}{1pt}{1pt}
Hyper                                                                         &  & \multicolumn{3}{c}{WeatherAUS}  &  &  & \multicolumn{3}{c}{Satimage}    &  \\ \cline{3-5} \cline{8-10}
Param.&  & Setting    &  & Weighted F1   &  &  & Setting    &  & Weighted F1   &  \\ \specialrule{1pt}{1pt}{1pt}
\multirow{6}{*}{\rotatebox[origin=c]{0}{$T$}}                            &  & 10           &  &   0.679\footnotesize{±0.002}  &  &  & 10           &  & 0.841\footnotesize{±0.008} &  \\
                                                                         &  & 50          &  &   \textbf{0.680\footnotesize{±0.003}}  &  &  & 50          &  & \textbf{0.857\footnotesize{±0.006}} &  \\ 
                                                                         &  & 100          &  &   0.670\footnotesize{±0.001}  &  &  & 100          &  & 0.846\footnotesize{±0.004} &  \\ 
                                                                         &  & 150          &  &   0.660\footnotesize{±0.002}  &  &  & 150          &  & 0.847\footnotesize{±0.005}
 &  \\ 
                                                                         &  & 200          &  &   0.666\footnotesize{±0.002}  &  &  & 200          &  & 0.845\footnotesize{±0.002} &  \\      
                                                                         &  & 300          &  &   0.669\footnotesize{±0.002}  &  &  & 300          &  & 0.849\footnotesize{±0.004} &  \\ \hline
                                                                    
\multirow{5}{*}{\rotatebox[origin=c]{90}{$(\beta_{min}, \beta_{max})$}}  &  & (0.01, 5.0)  &  &   0.671\footnotesize{±0.000}  &  &  & (0.01, 1.0)  &  & 0.843\footnotesize{±0.002} &  \\
                                                                         &  & (0.01, 10.0) &  &   0.676\footnotesize{±0.001}  &  &  & (0.01, 10.0) &  & \textbf{0.848\footnotesize{±0.003} }&  \\
                                                                         &  & (0.1, 1.0)   &  &   \textbf{0.676\footnotesize{±0.002}}  &  &  & (0.1, 1.0)   &  & 0.820\footnotesize{±0.004} &  \\
                                                                         &  & (0.1, 5.0)   &  &   0.673\footnotesize{±0.002}  &  &  & (0.1, 5.0)   &  & 0.834\footnotesize{±0.008} &  \\
                                                                         &  & (0.1, 10.0)  &  &   0.668\footnotesize{±0.003}  &  &  & (0.1, 10.0)  &  & 0.834\footnotesize{±0.014} &  \\ \hline
\multirow{2}{*}{\rotatebox[origin=c]{0}{$\xi$}}                          &  & 80           &  &   0.679\footnotesize{±0.003}  &  &  & 70           &  & 0.783\footnotesize{±0.016} &  \\
                                                                         &  & 90           &  &   \textbf{0.680\footnotesize{±0.002} } &  &  & 90           &  & \textbf{0.854\footnotesize{±0.005} }&  \\ \hline
\multirow{2}{*}{\rotatebox[origin=c]{0}{$w$}}                            &  & 0.99         &  &   0.680\footnotesize{±0.003}  &  &  & 0.99         &  & \textbf{0.856\footnotesize{±0.006}} &  \\
                                                                         &  & 0.90          &  &  \textbf{0.680\footnotesize{±0.003}} &  &  & 0.90          &  & 0.854\footnotesize{±0.006} &  \\ \hline
\multirow{4}{*}{\rotatebox[origin=c]{0}{$\epsilon_t$}}                   &  & 5e-04        &  & \textbf{0.680\footnotesize{±0.003}}  &  &  & 5e-04         &  & \textbf{0.857\footnotesize{±0.006}} &  \\
                                                                         &  & 1            &  & 0.661\footnotesize{±0.002}  &  &  & 1          &  & 0.850\footnotesize{±0.004} &  \\
                                                                         &  & 2            &  & 0.661\footnotesize{±0.002}  &  &  & 2          &  & 0.857\footnotesize{±0.004} &  \\
                                                                         &  & 3            &  & 0.661\footnotesize{±0.003}  &  &  & 3          &  & 0.854\footnotesize{±0.002} & \\
\specialrule{1pt}{1pt}{1pt}
\end{tabular}
\vspace{-0.5em}
\end{table}

\vspace{-0.5em}
\subsection{Sensitivity on Boundary vs. Regular Oversampling}
The first option of our sampling strategy (cf. Fig.~\ref{fig:patrick} (a)) corresponds to oversampling around class boundary, and the second option (cf. Fig.~\ref{fig:patrick} (b)) corresponds to regular oversampling. We compare their key results in Table~\ref{table:ZvsN}. It looks like that there does not exist a clear winner, but the boundary oversampling produces better results in more cases.

\vspace{-0.5em}
\subsection{Ablation study on the reverse SDE solver (or predictor)}
To solve $\hat{\mathbf{x}}^-_0$ from $\mathbf{x}^+_T$ with Eq.~\eqref{eq:reverse}, we can adopt various strategies. In general, the Euler-Maruyama (EM) method~\cite{platen_1999} is popular in solving SDEs. However, one can choose different SDE solvers, such as ancestral sampling~\cite{NEURIPS2020_4c5bcfec}, reverse diffusion~\cite{songyang}, probability flow~\cite{songyang} and so on. Following the naming convection of~\cite{songyang}, we call them as \emph{predictor} (instead of solver) in this subsection. In the context of SGMs, therefore, the predictor means the SDE solver to solve Eq.~\eqref{eq:reverse}.

In Table~\ref{table:pc_sampler}, the Euler-Maruyama method leads to the best outcomes for \texttt{Surgical}. For \texttt{Shoppers}, the ancestral sampling method produces the best outcomes. Likewise, predictors depend on datasets. The probability flow also shows reasonable results in our experiments. Especially, we observe that the probability flow always outperforms other predictors in \texttt{Buddy}.

\begin{table*}[]
\caption{Comparison between the boundary (Bnd.) vs. regular (Reg.) oversampling}
\label{table:ZvsN}
\small
\centering
\setlength{\tabcolsep}{3pt}
\begin{tabular}{c|cccc|cccc|cccc}
\specialrule{1pt}{1pt}{1pt}

SDE & &\multicolumn{2}{c}{WeatherAUS}  && &\multicolumn{2}{c}{Default} &&& \multicolumn{2}{c}{Satimage} & \\ \cline{3-4}\cline{7-8}\cline{11-12}
Type   && Bnd.           & Reg.          &&& Bnd.         & Reg.         &&& Bnd.          & Reg.          &\\ \specialrule{1pt}{1pt}{1pt}
\texttt{SOS} (VE)     && 0.670\footnotesize{±0.002}    & 0.672\footnotesize{±0.002}   &&& 0.569\footnotesize{±0.001}  & 0.571\footnotesize{±0.002}  &&& 0.854\footnotesize{±0.004}   & 0.854\footnotesize{±0.002}   &\\
\texttt{SOS} (VP)     && 0.673\footnotesize{±0.002}    & 0.680\footnotesize{±0.002}   &&& 0.559\footnotesize{±0.005}  & 0.557\footnotesize{±0.004}  &&& 0.855\footnotesize{±0.005}   & \textbf{0.857\footnotesize{±0.006}}  &\\
\texttt{SOS} (Sub-VP) && \textbf{0.680\footnotesize{±0.003}}    & 0.651\footnotesize{±0.002}   &&& \textbf{0.574\footnotesize{±0.003}}  & 0.574\footnotesize{±0.002}  &&& 0.854\footnotesize{±0.005}   & 0.855\footnotesize{±0.004}   & \\ 
\specialrule{1pt}{1pt}{1pt}

\end{tabular}
\end{table*}

\begin{table}
\small
\centering
\setlength{\tabcolsep}{2pt}

\caption{Results by the reverse SDE solver (or predictor). We consider the predictors of Euler-Maruyama (EM), Ancestral Sampling (AS), Reverse Diffusion (RD), and Probability Flow (PF).}\label{table:pc_sampler}
    \subtable[Surgical]{
        \centering
        \begin{tabular}{ccc|cccc|cccc}
        \specialrule{1pt}{1pt}{1pt}
        & \multirow{2}{*}{Predictor} &  & & \multicolumn{2}{c}{VE} & & & \multicolumn{2}{c}{VP} & \\ \cline{5-6} \cline{9-10}
        &  & & &   Pred. only  &  Pred. Corr.  & & &    Pred. only   &  Pred. Corr. & \\ 
        
        \specialrule{1pt}{1pt}{1pt}
        & EM          & &  & 0.709\footnotesize{±0.002} & \textbf{0.709\footnotesize{±0.003}} &     & & 0.710\footnotesize{±0.004} & \textbf{0.712\footnotesize{±0.002}} & \\
        & AS          & &  & 0.706\footnotesize{±0.002}  &  0.707\footnotesize{±0.003}  &    & & 0.706\footnotesize{±0.004} & 0.711\footnotesize{±0.002} & \\
        & RD          & & & 0.701\footnotesize{±0.003} &  0.708\footnotesize{±0.003}  &    & & 0.707\footnotesize{±0.002} &  0.711\footnotesize{±0.001} & \\
        & PF          & &  &  \multicolumn{2}{c}{0.704\footnotesize{±0.004}} &    & &  \multicolumn{2}{c}{0.712\footnotesize{±0.002}} & \\
        \specialrule{1pt}{1pt}{1pt}
        \end{tabular}
        \label{table:surgical}}
    \subtable[Shoppers]{
        \centering
        \begin{tabular}{ccc|cccc|cccc}
        \specialrule{1pt}{1pt}{1pt}
        & \multirow{2}{*}{Predictor} &  & & \multicolumn{2}{c}{VE} & & & \multicolumn{2}{c}{VP} & \\ \cline{5-6} \cline{9-10}
        &  & & &   Pred. only  &  Pred. Corr.  & & &    Pred. only   &  Pred. Corr. & \\ 
        
        \specialrule{1pt}{1pt}{1pt}
        & EM          & &  & 0.668\footnotesize{±0.003} & 0.664\footnotesize{±0.005}  &     & & 0.648\footnotesize{±0.007} & 0.651\footnotesize{±0.005} & \\
        & AS          & &  & \textbf{0.675\footnotesize{±0.006}} & 0.661\footnotesize{±0.003}  &    & & 0.646\footnotesize{±0.002}  & \textbf{0.657\footnotesize{±0.002}} & \\
        & RD          & &  & 0.673\footnotesize{±0.005} & 0.671\footnotesize{±0.006}  &    & & 0.640\footnotesize{±0.004}  & 0.655\footnotesize{±0.004} & \\
        & PF          & &  & \multicolumn{2}{c}{0.674\footnotesize{±0.004}}  &    & &  \multicolumn{2}{c}{0.640\footnotesize{±0.005}} & \\
        \specialrule{1pt}{1pt}{1pt}
        \end{tabular}
        \label{table:shoppers}}
\end{table}

\subsection{Ablation study on Adversarial Score Matching vs. Fine-tuning}
We also compare the two enhancement strategies, the adversarial score matching (cf. Fig.~\ref{fig:asm}) and our fine-tuning, in Table~\ref{table:abl_adv}. First of all, we found that the adversarial score matching for tabular data is not as effective as that for image data. The original method of the adversarial score matching had been designed for images to integrate SGMs with GANs. We simply took the discriminator of \texttt{CTGAN}, one of the best performing GAN models for tabular data, and combine it with our model, when building the adversarial score matching for our tabular data. Although there can be a better design, the current implementation with the best existing discriminator model for tabular data does not show good performance in our experiments. In many cases, it is inferior to our method even without fine-tuning. However, our fine-tuning successfully improves in almost all cases.

\vspace{-0.5em}

\begin{table*}[]
\caption{Results of No Fine-tuning vs. Adversarial SGM vs. Fine-tuning}
\label{table:abl_adv}
\small
\centering
\setlength{\tabcolsep}{2pt}
\begin{tabular}{c|ccccc|ccccc|ccccc}
\specialrule{1pt}{1pt}{1pt}

SDE 
&&   \multicolumn{3}{c}{\texttt{Shoppers}}  &
&&   \multicolumn{3}{c}{\texttt{Surgical}}  &
&&   \multicolumn{3}{c}{\texttt{Default}}  &\\  \cline{3-5} \cline{8-10} \cline{13-15}
Type
&&   No Fine-Tune    &   Adv. SGM    &   Fine-Tune &
&&   No Fine-Tune    &   Adv. SGM    &   Fine-Tune &
&&   No Fine-Tune    &   Adv. SGM    &   Fine-Tune &\\ \specialrule{1pt}{1pt}{1pt}
\texttt{SOS} (VE)  
&&  0.672\footnotesize{±0.003}  &   0.651\footnotesize{±0.005}    &   \textbf{0.675\footnotesize{±0.004}}&
&&  0.709\footnotesize{±0.004} &   0.692\footnotesize{±0.003} &   0.709\footnotesize{±0.003}	&
&&	0.567\footnotesize{±0.001}	&	0.565\footnotesize{±0.002}	&	0.571\footnotesize{±0.002}   &	\\
\texttt{SOS} (VP)	 
&&	0.652\footnotesize{±0.004}	&	0.651\footnotesize{±0.005}	&	0.658\footnotesize{±0.002}             &
&&	0.708\footnotesize{±0.002} &	0.689\footnotesize{±0.001} 	&	0.712\footnotesize{±0.002}   &
&&	0.558\footnotesize{±0.004}	&	0.557\footnotesize{±0.004}	&	0.559\footnotesize{±0.006}   &	\\
\texttt{SOS} (Sub-VP)
&&	0.667\footnotesize{±0.005}	&	0.659\footnotesize{±0.003}	&	0.673\footnotesize{±0.002}   &
&&	0.712\footnotesize{±0.003}	&	0.705\footnotesize{±0.002}	&	\textbf{0.714\footnotesize{±0.001}}   &
&&	0.570\footnotesize{±0.001}	&	0.555\footnotesize{±0.003}	&	\textbf{0.574\footnotesize{±0.003}}   &   \\ 
\specialrule{1pt}{1pt}{1pt}

\end{tabular}
\vspace{-5pt}
\end{table*}






\subsection{Visualization}
We introduce several key visualization results, ranging from histograms of columns to t-SNE plots.
\vspace{-5pt}
\subsubsection{Column-wise Histogram}
\begin{figure}[t]
    \centering
    \subfigure[\texttt{Surgical} (bmi)]{\includegraphics[width=0.48\columnwidth]{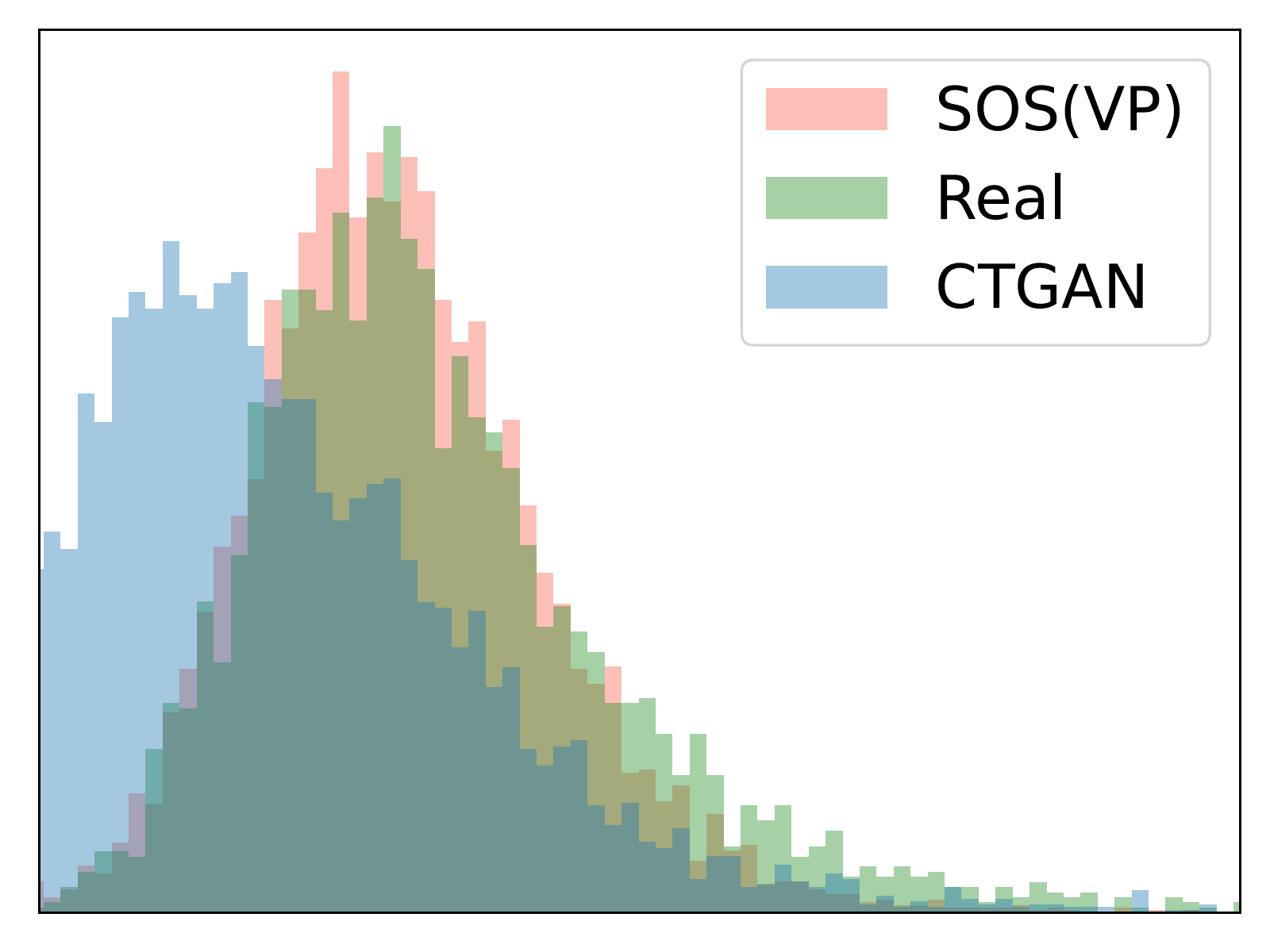}}
    \subfigure[\texttt{WeatherAUS} (temperature)]{\includegraphics[width=0.48\columnwidth]{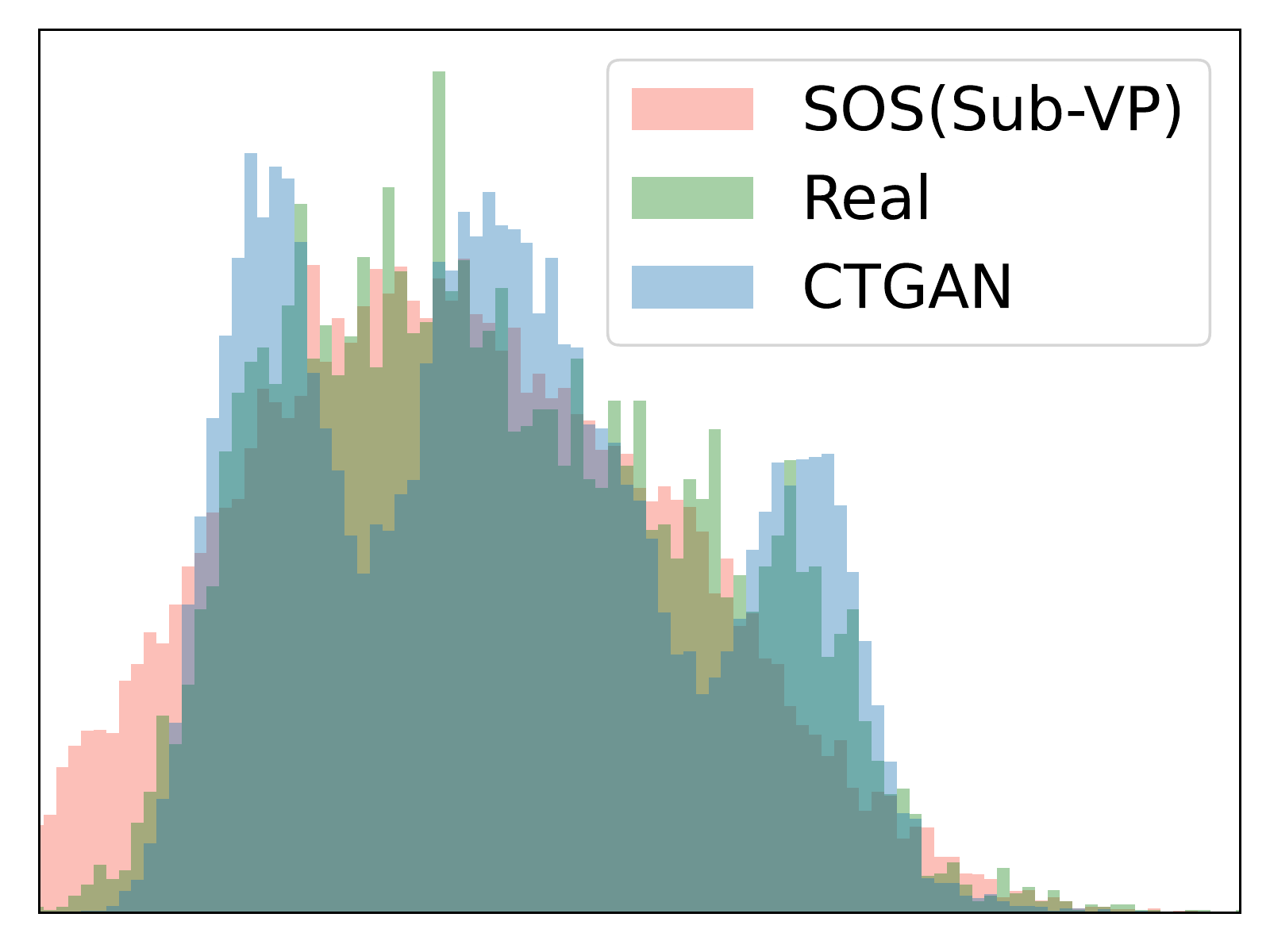}}
    \caption{Column-wise Histogram}
    \label{fig:histogram}
    \vspace{-1em}
\end{figure}

Fig.~\ref{fig:histogram} shows two histogram figures. The real histogram of values and the histogram by our method are similar to each other in both figures. However, \texttt{CTGAN} fails to capture the real distribution of the `bmi' column of \texttt{Surgical} in Fig.~\ref{fig:histogram} (a), which we consider the well-known mode-collapse problem of GANs, i.e., values in certain regions are more actively sampled than those in other regions. In Fig.~\ref{fig:histogram} (b), \texttt{CTGAN}'s histogram shows three peaks whereas the real histogram has only one peak. Our method successfully captures it.

\begin{figure}[t]
    \centering
    \includegraphics[width=0.8\columnwidth]{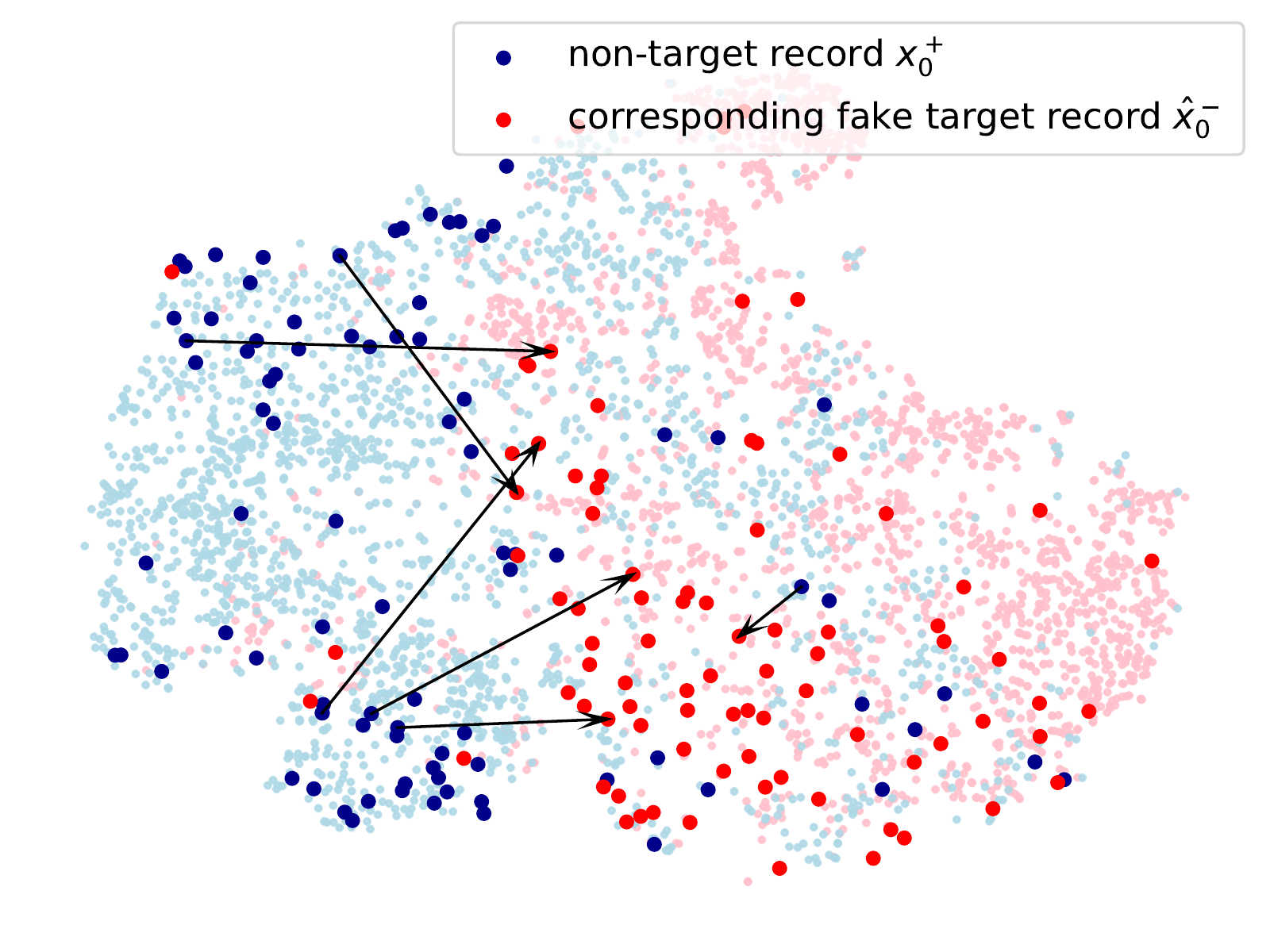}
    \caption{t-SNE plot of real/fake records by our method with boundary oversampling on \texttt{WeatherAUS}. Blue dots mean non-target records, i.e., $\mathbf{x}^+_0$, and red dots mean fake target records, i.e., $\hat{\mathbf{x}}^-_0$. Black arrows show which non-target records are style-transferred to which target records.}
    \label{fig:majortominor}
    \vspace{-1em}
\end{figure}

\subsubsection{t-SNE plot of fake/real records}
In Fig.~\ref{fig:majortominor}, we visualize real and fake records --- we use \texttt{SOS} (Sub-VP) with the boundary oversampling option {in \texttt{WeatherAUS}}. The translucent dots mean real records, and the solid blue dots are transferred to the solid red dots in the figure. As described earlier, those solid red dots are around the class boundary.

\section{Conclusions \& Limitations}
Oversampling minor classes is a long-standing research problem in data mining and machine learning. Many different methods have been proposed so far. In this paper, however, we presented a score-based oversampling method, called \texttt{SOS}. Since SGMs were originally invented for images, we carefully redesigned many parts of them. For instance, we designed a neural network to approximate the score function of the forward SDE and a fine-tuning method. In our experiments with 6 datasets and 10 baselines, our method overwhelms other oversampling methods in all cases. Moreover, our method does not decrease the F1 score after oversampling in all cases whereas existing methods fail to do so from time to time. From all these facts, we think that our method makes a great step forward in oversampling minor classes.

Even though SGMs perform well in general, they are still in an early phase when it comes to generating tabular data, and we hope that our work will bring much follow-up research. For instance, one can invent a drift and/or a diffusion function different from the three existing types and specialized to tabular data.

\begin{acks}
Jayoung Kim and Chaejeong Lee equally contributed. Noseong Park is the corresponding author. This work was supported by the Institute of Information \& Communications Technology Planning \& Evaluation (IITP) grant funded by the Korea government (MSIT) (10\% from No. 2020-0-01361, Artificial Intelligence Graduate School Program at Yonsei University and 70\% from No. 2021-0-00231, Development of Approximate DBMS Query Technology to Facilitate Fast Query Processing for Exploratory Data Analysis) and the National Research Foundation of Korea (NRF) grant funded by the Korea government (MSIT) (20\% from No. 2021R1F1A1063981).

\end{acks}

\bibliographystyle{ACM-Reference-Format}
\bibliography{ref}

\clearpage
\appendix
\section{Settings \& Reproducibility}\label{a:hyper}
Our software and hardware environments are as follows: \textsc{Ubuntu} 18.04 LTS, \textsc{Python} 3.8.2, \textsc{Pytorch} 1.8.1, \textsc{CUDA} 11.4, and \textsc{NVIDIA} Driver 470.42.01, i9 CPU, and \textsc{NVIDIA RTX 3090}. Our codes and data are at \url{https://github.com/JayoungKim408/SOS}.

In Tables~\ref{table:hyperparameters_sgm} and~\ref{table:hyperparameters_finetune}, we list the best hyperparameters. We have three layer types as shown in Appendix~\ref{a:archi}: Concat, Squash, and Concatsquash~\cite{grathwohl2018ffjord}. $(\beta_{min}, \beta_{max})$ (resp. $(\sigma_{min}, \sigma_{max})$) are two hyperparameters of the function $\beta(t)$ (resp. $\sigma(t)$). In fact, the full writing of $\beta$ and $\sigma$ are $\beta(t;\beta_{min}, \beta_{max})$ and $\sigma(t;\sigma_{min}, \sigma_{max})$, respectively. We search for (min, max), in total, with 9 combinations using $min = \{0.01, 0.1, 0.5\}$ and $max = \{1.0, 5.0, 10.0\}$. We use a learning rate in \{$1 \times 10^{-i}$, $2 \times 10^{-i}$ | $i$ = \{3, 4, 5\}\}. Our predictors are in \{AS, RD, EM, PF\} and the corrector is the Langevin corrector~\cite{songyang}. SNR is the signal-to-noise ratio set to $\{0.01, 0.05, 0.16\}$ when using the corrector.

We also consider the following hyperparameter configurations for the fine-tuning process: the fine-tuning learning rate is \{$2 \times 10^{-i}$ | $i$ = \{4, 5, \dots, 8\}\}, and the angle threshold $\xi$ is \{$60+5 \times i$ | $i$ = \{0, 1, \dots, 10\}\}. The number of fine-tuning epochs is \{1, 2, 3, 4, 5\}, and $w$ is \{0.99, 0.95, 0.9, 0.8, 0.7, 0.6\}. $\epsilon_{t}$ is \{5e-04, 1, 2, 3\} in all datasets.

\begin{table*}[t]
\caption{The best hyperparameters of \texttt{SOS} we used in Table~\ref{table:main}. `(min, max)' means $(\beta_{min}, \beta_{max})$ for VP and Sub-VP or $(\sigma_{min}, \sigma_{max})$ for VE, respectively.}
\label{table:hyperparameters_sgm}
\small
\centering
\setlength{\tabcolsep}{2pt}
\begin{tabular}{c|c|cccccccccc}

\specialrule{1pt}{1pt}{1pt}

\multirow{2}{*}{Dataset} 
& SDE &
&   \multicolumn{8}{c}{Hyperparameters for \texttt{SOS}}  &\\  
\cline{4-11}

& Type &
& Layer Type & $( \dim(h_1), \dim(h_1), \dots, \dim(h_{d_{N}}))$ &  Activation & Learn. Rate  &   (min, max)  &  Pred. & Corr. & SNR &\\ 
\specialrule{1pt}{1pt}{1pt}
\multirow{3}{*}{\texttt{Default}}
& VE &
&\multirow{3}{*}{Concat}  &	(512, 1024, 1024, 512)	& \multirow{3}{*}{LeakyReLU} 
& \multirow{3}{*}{2e-03}  & (0.01, 5.0) & RD & Langevin & 0.05  &  \\
& VP &
&&(256, 512, 1024, 1024, 512, 256)&
& - & (0.1, 5.0) & PF & None & -  & \\
& Sub-VP &
&&(512, 1024, 1024, 512)& 
&  & (0.01, 1.0) & EM & None &  - & \\
\hline
\multirow{3}{*}{\texttt{Shoppers}}
& VE &
&	\multirow{3}{*}{Concatsquash}   &	\multirow{3}{*}{(512, 1024, 2048, 1024, 512)}	& \multirow{3}{*}{ReLU} 
& \multirow{3}{*}{1e-03} & (0.1, 5.0) & AS & None & - &  \\
& VP &
&&&  
& - & (0.01, 5.0) & RD & Langevin & 0.05 & \\
& Sub-VP &
&&& 
&  & (0.1, 10.0) & EM & None &  -  & \\
\hline
\multirow{3}{*}{\texttt{Surgical}}
& VE &
&	\multirow{2}{*}{Squash}   &	\multirow{2}{*}{(256, 512, 1024, 1024, 512, 256)}	&	 \multirow{3}{*}{ReLU} & \multirow{3}{*}{2e-03} & (0.1, 10.0)  & EM & Langevin & 0.05  &  \\
& VP & &&&   &  & (0.01, 5.0) & EM & Langevin & 0.16  & \\
& Sub-VP &
& Concat & (512, 1024, 1024, 512) & &  & (0.1, 10.0)  & PF & None &  -  & \\
\hline
\multirow{3}{*}{\texttt{WeatherAUS}}
& VE &
&	\multirow{3}{*}{Concat}   &	\multirow{3}{*}{(512, 1024, 2048, 1024, 512)}	&	 \multirow{3}{*}{LeakyReLU} 
& \multirow{2}{*}{2e-04} & (0.01, 5.0) & AS & Langevin & 0.16  &  \\
& VP &
&&&  
&  & (0.01, 10.0) & RD & None & -  & \\
& Sub-VP &
&&& 
& 2e-03 & (0.01, 1.0) & EM & None &  -  & \\
\hline
\multirow{3}{*}{\texttt{Buddy}}
& VE &
&	Concat   &\multirow{3}{*}{(256, 512, 1024, 1024, 512, 256)}& \multirow{3}{*}{SoftPlus} & \multirow{3}{*}{2e-03} & (0.5, 5.0)&PF  & None & - &\\
& VP &
& \multirow{2}{*}{Squash} & &  
&  & (0.5, 10.0) & PF & None & -  & \\
& Sub-VP &
& & &  
&  & (0.1, 10.0) & PF & None &  -  & \\
\hline
\multirow{3}{*}{\texttt{Satimage}}
& VE &
&	\multirow{3}{*}{Concat}   &	\multirow{3}{*}{(512, 1024, 2048, 2048, 1024, 512)}	&	 \multirow{3}{*}{LeakyReLU} 
& \multirow{3}{*}{2e-04}  & (0.01, 5.0) & RD & None & -  &  \\
& VP &
&&&  
& - & (0.01, 5.0) & RD & Langevin & 0.05  & \\
& Sub-VP &
&&& 
&  & (0.1, 5.0) & EM & None &  - & \\
\specialrule{1pt}{1pt}{1pt}
\end{tabular}

\end{table*}

\begin{table}[]
\caption{The best hyperparameters of the fine-tuning process we used in Table~\ref{table:main} }
\label{table:hyperparameters_finetune}
\small
\centering
\setlength{\tabcolsep}{3pt}
\begin{tabular}{c|c|cccccccc}

\specialrule{1pt}{1pt}{1pt}

\multirow{2}{*}{Dataset} 
& SDE &
&   \multicolumn{6}{c}{Hyperparameters for fine-tuning} &\\  
\cline{4-9}
 
& Type &
& $\epsilon_{t}$ & Learn. Rate  &    $\xi$  & $w$ & Epoch & Option & \\ 
\specialrule{1pt}{1pt}{1pt}
\multirow{3}{*}{\texttt{Default}}
& VE &
& 5e-04 &	2e-06	&	80	& 0.90 & 4 & Reg. & \\
& VP &
& 2 &	2e-08	&	80	&	0.95 & 1 & Bnd. &\\
& Sub-VP &
& 5e-04 &	2e-06	&	90	&	0.95 & 1 &  Bnd. &\\
\hline
\multirow{3}{*}{\texttt{Shoppers}}
& VE &
& 5e-04 &	2e-07	&	90	& 0.80 & 1 & Bnd. &\\
& VP &
&  5e-04 &	2e-06	&	100	&	0.95 & 3 & Reg. &\\
& Sub-VP &
& 5e-04 &	2e-05	&	80	&	0.95	& 2 & Bnd.&\\
\hline
\multirow{3}{*}{\texttt{Surgical}}
& VE &
& 5e-04&	2e-08	&	80	& 0.99 & 3 & Reg. &\\
& VP &
& 5e-04 &	2e-06	&	100	&	0.90 & 5 & Reg. &\\
& Sub-VP &
& 2 &	2e-07	&	80	&	0.95	& 3 & Bnd.&\\
\hline
\multirow{3}{*}{\texttt{WeatherAUS}}
& VE &
& 5e-04 &	2e-05	&	100	& 0.70 & 4 & Reg. &\\
& VP &
& 5e-04 &	2e-07	&	60	&	0.60 & 1 & Reg. &\\
& Sub-VP &
& 5e-04 &	2e-07	&	100	&	0.95	& 1 & Bnd. &\\
\hline
\multirow{3}{*}{\texttt{Buddy}}
& VE &
& 5e-04 &	2e-07	&	80	& 0.99 & 3 & Bnd. &\\
& VP &
& 5e-04 &	2e-07 &	80	& 0.90 & 3 & Bnd.&\\
& Sub-VP &
& 2 &	2e-03	&	80	&	0.90	& 4 & Bnd. &\\
\hline
\multirow{3}{*}{\texttt{Satimage}}
& VE &
& 5e-04 &	2e-07	&	80	& 0.95 & 3 & Reg. &\\
& VP &
& 5e-04 &	2e-06	&	80	&	0.95 & 4 & Reg. &\\
& Sub-VP &
& 5e-04 &	2e-08	&	80	&	0.70	& 5 & Reg. & \\
\specialrule{1pt}{1pt}{1pt}
\end{tabular}

\end{table}


\begin{table*}[t]
\small
\centering
\setlength{\tabcolsep}{3pt}
\caption{Results of experiments synthesizing full fake tabular data}\label{tbl:entire_synthesis}
\begin{tabular}{ccc|cccc|cccc|cccc}
\specialrule{1pt}{1pt}{1pt}
& \multirow{2}{*}{Method} &  & & \multicolumn{2}{c}{\texttt{Shoppers}} & & &  \multicolumn{2}{c}{\texttt{Surgical}}  & & &  \multicolumn{2}{c}{\texttt{Satimage}}  &    \\  \cline{5-6} \cline{9-10} \cline{13-14}

&  & &  & Acc. & F1 & & & Acc. & F1  & & & Acc. & F1 & \\  
\specialrule{1pt}{1pt}{1pt}

& \texttt{Identity}           & & & 0.883\footnotesize{±0.002}  & 0.503\footnotesize{±0.005}   & & & 0.827\footnotesize{±0.002} & 0.608\footnotesize{±0.005} & & & 0.846\footnotesize{±0.002}  & 0.799\footnotesize{±0.003}  &  \\
\hline
& \texttt{CTGAN}              & & & 0.778\footnotesize{±0.009}  & 0.440\footnotesize{±0.012}   & & & 0.716\footnotesize{±0.004} & 0.472\footnotesize{±0.003} & & &  0.735\footnotesize{±0.010}  & 0.699\footnotesize{±0.007}  &   \\
& \texttt{OCT-GAN}            & & & 0.835\footnotesize{±0.011}  & 0.490\footnotesize{±0.010}   & & & 0.695\footnotesize{±0.026} & 0.527\footnotesize{±0.014} & & &  0.798\footnotesize{±0.008}  & 0.767\footnotesize{±0.009}  &    \\
& \texttt{TableGAN}           & & & 0.854\footnotesize{±0.024}  & 0.515\footnotesize{±0.013}   & & & 0.752\footnotesize{±0.003} & 0.496\footnotesize{±0.009} & & &  0.783\footnotesize{±0.007}  & 0.710\footnotesize{±0.007}  &    \\
& \texttt{TVAE}               & & & 0.855\footnotesize{±0.004}  & 0.476\footnotesize{±0.009}   & & & 0.778\footnotesize{±0.004} & 0.427\footnotesize{±0.019} & & &  0.825\footnotesize{±0.004}  & 0.779\footnotesize{±0.008}  &    \\
\hline
& SOS          & & &  \textbf{0.874\footnotesize{±0.002}} & \textbf{0.618\footnotesize{±0.003}}  & & &  \textbf{0.798\footnotesize{±0.003}}  & \textbf{0.549\footnotesize{±0.005}}   & & &  \textbf{0.852\footnotesize{±0.003}} & \textbf{0.815\footnotesize{±0.004}} &  \\
\specialrule{1pt}{1pt}{1pt}
\end{tabular}
\end{table*}

\section{VE, VP, and Sub-VP SDEs}\label{sec:fandg}

The definitions of $f$ and $g$ as follows:
\begin{align}
    f(t)&=\begin{cases}0,  &\textrm{ if VE,}\\
    -\frac{1}{2}\beta(t)\mathbf{x},&\textrm{ if VP,}\\
    -\frac{1}{2}\beta(t)\mathbf{x},&\textrm{ if Sub-VP,}\\
    \end{cases}\\
    g(t)&=\begin{cases}\sqrt{\frac{\texttt{d}[\sigma^2(t)]}{\texttt{d}t}},&\textrm{ if VE,}\\
    \sqrt{\beta(t)},&\textrm{ if VP,}\\
    \sqrt{\beta(t)(1-e^{-2\int_0^t \beta(s)\, \texttt{d}s})},&\textrm{ if Sub-VP,}\\
    \end{cases}
\end{align}where {$\sigma(t)$ and $\beta(t)$ are noise functions at time $t$.}

\section{Score Network Architecture}\label{a:archi}

The proposed score network $S_{\theta}(\mathbf{x}, t)$ is as follows:
\begin{align*}
    \mathbf{h}_0     &=  \mathbf{x}_t,\\
    \mathbf{h}_i     &=  \omega( \mathtt{H}_i(\mathbf{h}_{i-1}, t) \oplus \mathbf{h}_{i-1} ), 1 \le i \le d_N\\
    S_{\boldsymbol{\theta}}(\mathbf{x}_t, t) &= \mathtt{FC}(\mathbf{h}_{d_N}),
\end{align*}where $\mathbf{x}_t$ is a record (or a row) at time $t$ in tabular data and $\omega$ is an activation function. $d_N$ is the number of hidden layers. For various layer types of $\mathtt{H}_i(\mathbf{h}_{i-1}, t)$, we provide the following options:
\begin{align*}
    \mathtt{H}_i(\mathbf{h}_{i-1}, t) &= 
        \begin{cases}
            \texttt{FC}_{i}(\mathbf{h}_{i-1}) \odot \psi(\texttt{FC}^{t}_{i}(t)), &\textrm{ if Squash,}\\
            \texttt{FC}_{i}(t \oplus \mathbf{h}_{i-1}), &\textrm{ if Concat,} \\
            \texttt{FC}_{i}(\mathbf{h}_{i-1}) \odot \psi(\texttt{FC}^{gate}_{i}(t) + \texttt{FC}^{bias}_{i}(t)), &\textrm{ if Concatsquash,}
        \end{cases}
\end{align*}where we can choose one of the three possible layer types as a hyperparameter, $\odot$ means the element-wise multiplication, $\oplus$ means the concatenation operator, $\psi$ is the Sigmoid function, and $\mathtt{FC}$ is a fully connected layer.

\section{Space and Time Overheads}\label{a:comp}

\begin{figure}[t]
    \centering
    \includegraphics[width=0.7\columnwidth]{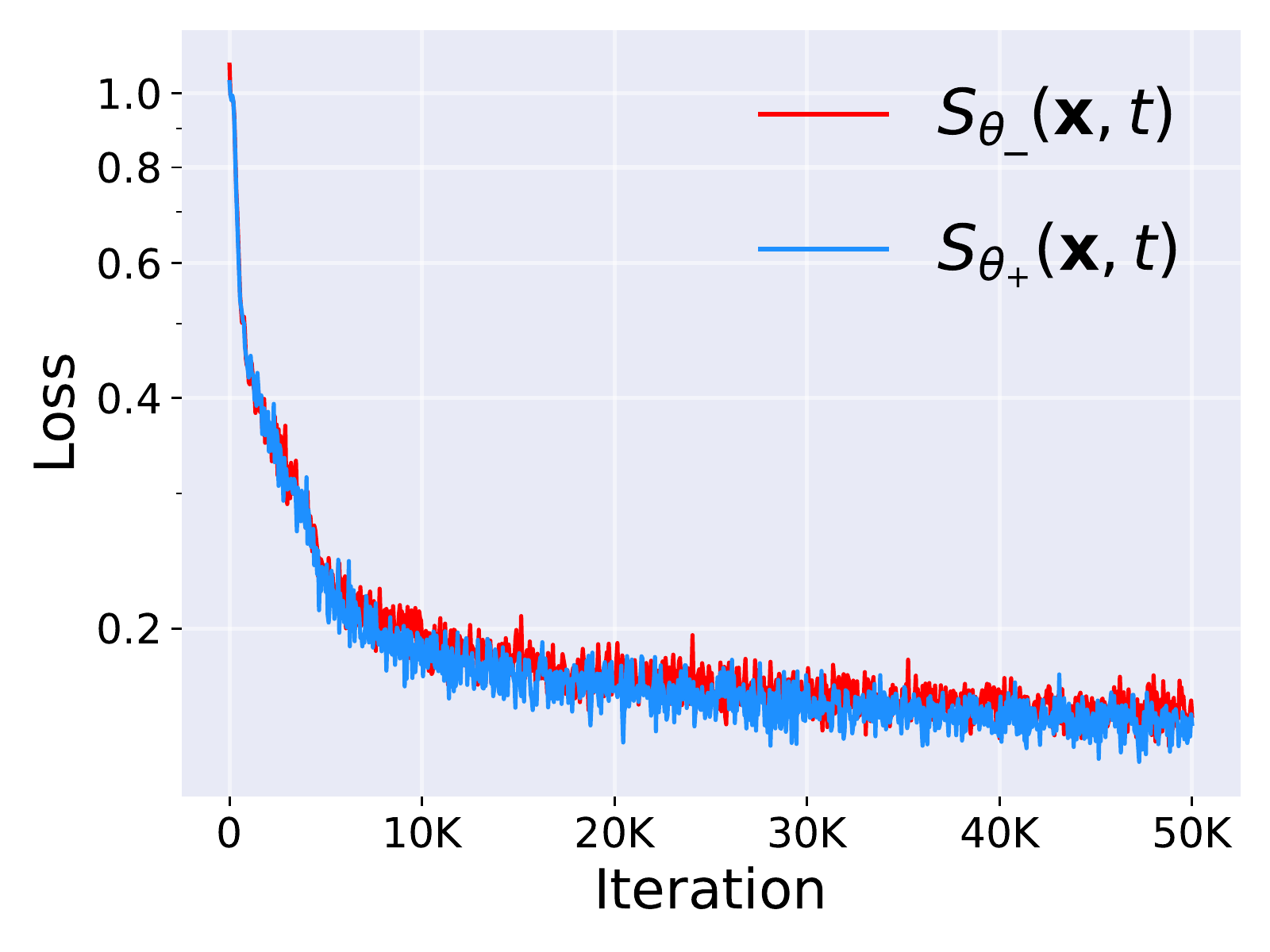}
    \caption{The training loss curves of \texttt{SOS} in \texttt{Default}}
    \label{fig:loss_curve}
\end{figure}

Some SGMs are notorious for their high computational overheads. In the case of tabular data, however, it is not the case since the number of columns is typically much smaller than other cases, e.g., the number of pixels of images. We introduce our training and synthesis overheads. As shown in Fig.~\ref{fig:loss_curve}, our training curves are smooth and in Table~\ref{table:time}, the space and time overheads of our method are well summarized. The GPU memory requirements are around 4GB, which is small. It takes well less than a second per epoch for training. For the total generation time, i.e, the time taken to accomplish the entire oversampling task, our method shows less than 5 seconds in all cases. Sometimes, it takes less than a second.

\begin{table}[t]
\caption{The runtime and GPU memory usage (averaged across classes) of \texttt{SOS} (Sub-VP) with the EM predictor.}
\label{table:time}
\small
\centering
\setlength{\tabcolsep}{2pt}



\begin{tabular}{c|c|c|c|c}
\specialrule{1pt}{1pt}{1pt}
\multirow{2}{*}{Dataset}    &   Avg.        &   GPU Memory    &   Train Time        &   Total Gen. \\ 
                            &   Class size  &   Usage (MB)    &   per Epoch (sec)   &   Time (sec) \\  \specialrule{1pt}{1pt}{1pt}
\texttt{Satimage}   &   3,186   &   4,424   &   0.036   &   4.240  \\
\texttt{Shoppers}   &   6,165   &   4,608   &   0.067   &   0.824 \\
\texttt{Surgical}   &   7,318   &   4,616   &   0.185   &   0.952 \\ 
\texttt{Buddy}      &   9,417   &   4,357   &   0.157   &   1.718 \\
\texttt{Default}    &   15,000  &   4,196   &   0.395   &   0.901 \\
\texttt{WeatherAUS} &   28,210  &   4,812   &   0.606   &   1.766 \\
\specialrule{1pt}{1pt}{1pt}
\end{tabular}
\end{table}

\begin{figure}[t]
    \centering
    \includegraphics[width=0.7\columnwidth]{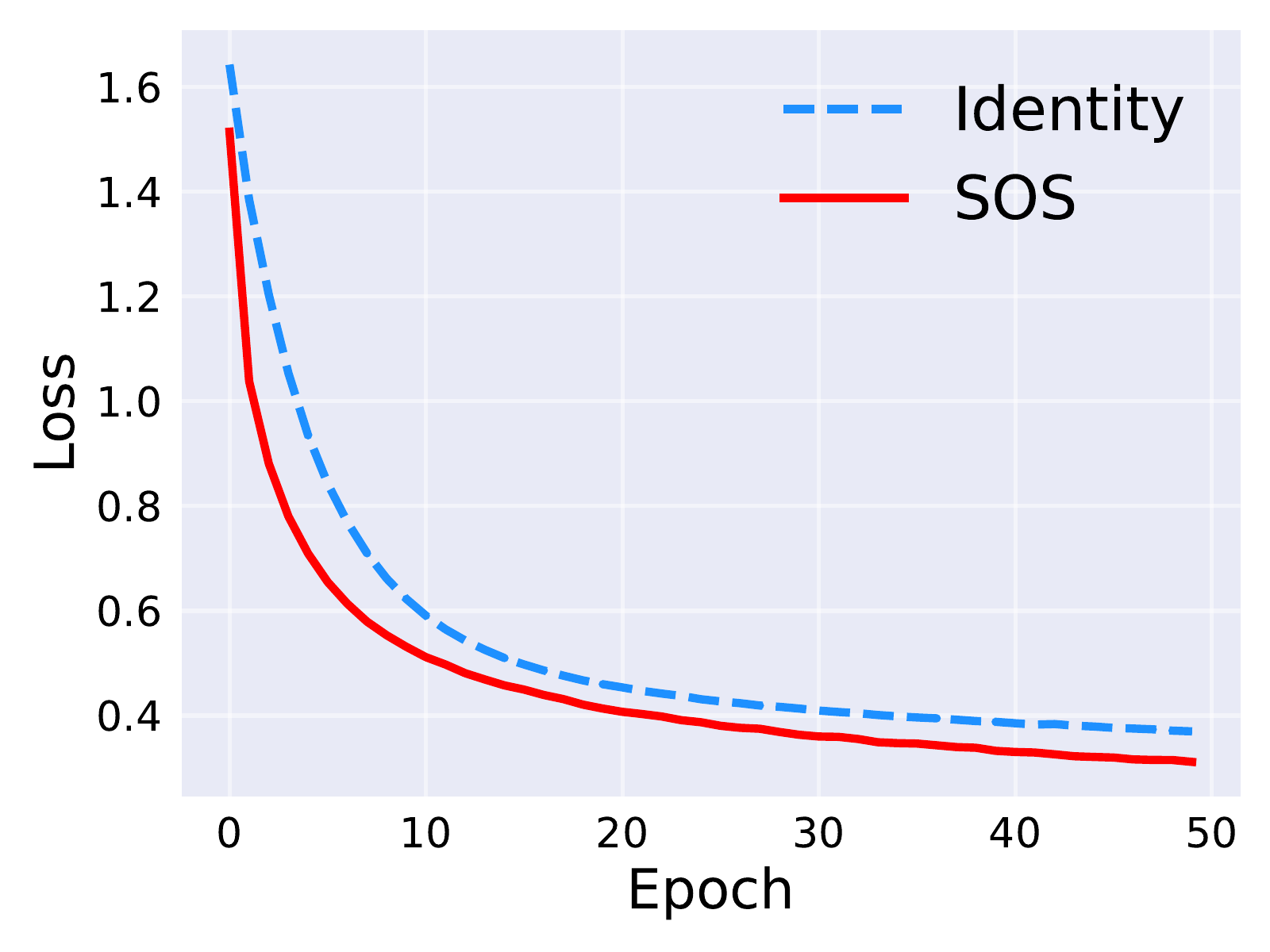}
    \caption{The training loss curves of the MLP classifier before and after oversampling in \texttt{Satimage}}
    \label{fig:full_fake_loss_curve}
\end{figure}

\section{Full Fake Table Synthesis}\label{a:entire}
Instead of oversampling minor classes, one can use our method for generating fake tabular data entirely --- in other words, fake tabular data consists of only fake records. To this end, we train one score network of \texttt{SOS} (Sub-VP) with all classes without the fine-tuning process since we do not distinguish major/minor classes but try to generate all classes. In Table~\ref{tbl:entire_synthesis}, we summarize its results in three datasets. As shown, it overwhelms all other existing methods. Even in comparison with \texttt{Identity}, it shows higher F1 scores in \texttt{Shoppers} and \texttt{Satimage}. However, its accuracy is lower than \texttt{Identity} in \texttt{Shoppers} and \texttt{Surgical}.

One interesting point is that for \texttt{Satimage}, our fake table shows better scores than \texttt{Identity} for all metrics. We compare the loss curves of the MLP classifier with the original data and our fake data in Fig.~\ref{fig:full_fake_loss_curve}. Surprisingly, the fake tabular data by \texttt{SOS} yields lower training loss values than \texttt{Identity}. For some reasons, the fake tabular data is likely to provide better training, which is very promising. We think that it needs more study on synthesizing fake tabular data with SGMs to fully understand the phenomena.

\end{document}